%% file: main.tex
\documentclass[letterpaper]{article} 
\usepackage{aaai2026}  
\usepackage{times}  
\usepackage{helvet}  
\usepackage{courier}  
\usepackage[hyphens]{url}  
\usepackage{graphicx} 
\usepackage{natbib}  
\usepackage{caption} 
\usepackage{algorithm}
\usepackage{algorithmic}
\usepackage{amsmath}
\usepackage{amsfonts} 
\usepackage{booktabs}
\usepackage{multirow}
\usepackage{rotating}

\usepackage{subfigure} 
\usepackage{newfloat}
\usepackage{listings}
\DeclareCaptionStyle{ruled}{labelfont=normalfont,labelsep=colon,strut=off} 
\floatstyle{ruled}
\newfloat{listing}{tb}{lst}{}
\floatname{listing}{Listing}
\title{FedSDWC: Federated Synergistic  \\ Dual-Representation Weak Causal Learning for OOD}
\author{
    Zhenyuan Huang \textsuperscript{\rm 1} \equalcontrib, 
    Hui Zhang \textsuperscript{\rm 1} \equalcontrib,
    Wenzhong Tang \textsuperscript{\rm 1} \equalcontrib,
    Haijun Yang \textsuperscript{\rm 1} \equalcontrib \thanks{Corresponding Author.}
}
\affiliations{
    \textsuperscript{\rm 1} Beihang University \\
    
    zhenyuan@buaa.edu.cn, hzhang@buaa.edu.cn, tangwenzhong@buaa.edu.cn, navy@buaa.edu.cn
}

\usepackage{bibentry}

\begin{document}

\maketitle

\begin{abstract}
Amid growing demands for data privacy and advances in computational infrastructure, federated learning (FL) has emerged as a prominent distributed learning paradigm. Nevertheless, differences in data distribution (such as covariate and semantic shifts) severely affect its reliability in real-world deployments. To address this issue, we propose FedSDWC, a causal inference method that integrates both invariant and variant features. FedSDWC infers causal semantic representations by modeling the weak causal influence between invariant and variant features, effectively overcoming the limitations of existing invariant learning methods in accurately capturing invariant features and directly constructing causal representations.
This approach significantly enhances FL's ability to generalize and detect OOD data.  
Theoretically, we derive FedSDWC's generalization error bound under specific conditions and, for the first time, establish its relationship with client prior distributions. 
Moreover, extensive experiments conducted on multiple benchmark datasets validate the superior performance of FedSDWC in handling covariate and semantic shifts. 
For example, FedSDWC outperforms FedICON, the next best baseline, by an average of 3.04\% on CIFAR-10 and 8.11\% on CIFAR-100.
\end{abstract}


\section{Introduction}\label{sec:intro}
\begin{figure}[t]
\centering
\includegraphics[width=1.0\columnwidth]{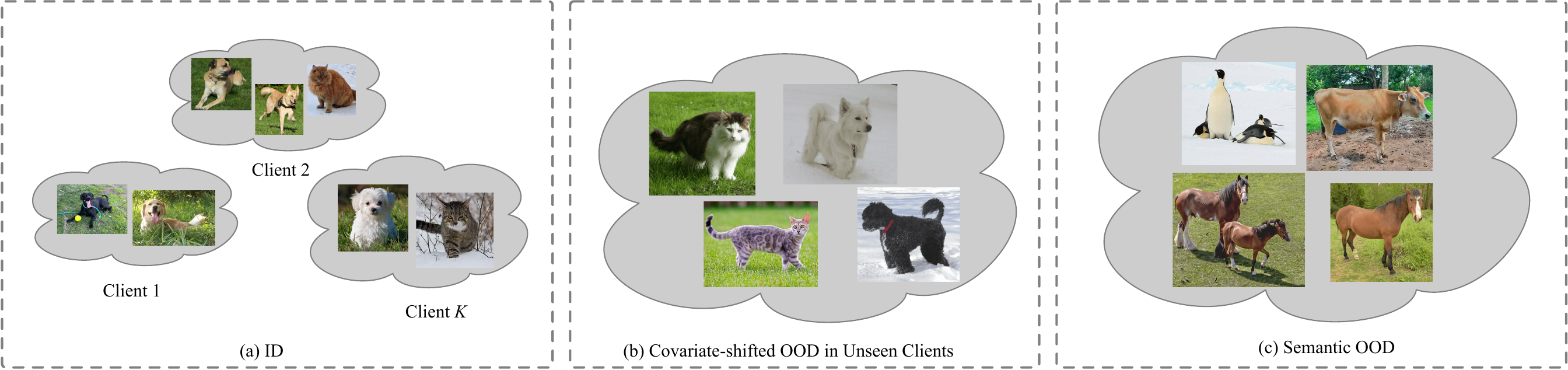} 
\caption{
Taking the task of identifying cats and dogs as an example, FL faces three major data challenges in the real world. (1) \textbf{In-Distribution (ID) Data} refers to the training data from participating clients. This data is often heterogeneous; for example, different clients might have images of dogs with a grass background or cats with a snow background, leading to non-identical class or feature distributions across the overall training data.
(2) \textbf{Covariate-Shifted OOD Data} refers to data from non-participating clients or data from participating clients that was not used for training, where the feature distribution has changed. For example, the dog’s background shifts from grass to snow, and the cat’s background shifts from snow to grass.
(3) \textbf{Semantic-Shifted OOD Data} refers to the emergence of new categories not present in the training set, such as cows and horses.
}
\label{icml-historical}
\end{figure}

Federated Learning (FL) has emerged as a key distributed learning paradigm for its ability to enable collaborative model training while preserving data privacy \cite{mcmahan2017communication, liao2024foogd}. However, its practical application is hindered by significant challenges, most notably the non-independent and identically distributed (non-IID) nature of client data \cite{li2020federated}.
Beyond data heterogeneity, a more pressing challenge is out-of-distribution (OOD) generalization \cite{jiang2022test, sefidgaran2024lessons, qi2025federated}. 
In real-world FL, training occurs on a subset of clients and their data, creating a distributional shift—known as covariate shift—between the training data and the true data population (as shown in Figure \ref{icml-historical}(b)). 
This phenomenon is known as the OOD generalization problem in FL \citep{liao2024foogd, zhou2025fedgog}, where the core task is to capture stable feature-label relationships under covariate shift, enabling the model to generalize effectively to unseen clients.
Under this context, FL must also address the semantic shift problem, which involves identifying data samples that do not belong to known categories during training. For example, in Figure \ref{icml-historical}(c), categories such as cows and horses have not been encountered during training. Therefore, the model should be capable of rejecting such data rather than misclassifying them into known categories.
Concurrently, FL models must handle semantic shifts, where they need to perform OOD detection to identify and reject data from new categories not seen during training (e.g., identifying a cow when trained only on cats and dogs in Fig. \ref{icml-historical}(c)), rather than misclassifying them.

Existing research on OOD generalization, including disentangled learning \citep{lee2023diversify, wang2023generalized, bai2024diprompt}, invariant risk minimization (IRM) \citep{arjovsky2019invariant, li2022learning, guo2023out}, and causal inference \citep{liu2021learning, bravo2024intervention, noohdani2024decompose, zhang2025federated} primarily aims to extract invariant features that are stable across different data distributions. 
However, these methods are often limited by the restrictive assumption that a representation function \(\Psi(\cdot)\) exists, which can extract features from the input data \(X\) that are perfectly invariant across any two environments \((e,{e'})\), such that \(\Psi(X^e) = \Psi(X^{e'})\).
This assumption is difficult to satisfy in practice. Furthermore, by focusing exclusively on invariant features, these approaches risk discarding valuable information contained within the variant features of the data. Directly extracting causal features from complex data like images also remains a significant hurdle.

To address these limitations, 
we relax the strict invariance assumption, instead assuming that our feature mapping can extract most, but not all, invariant information. Crucially, we incorporate the often-overlooked variant features, positing they contain a smaller but still valuable amount of information. Based on this, we construct a causal graph (Fig. \ref{icml-historical}) between the latent factors derived from both invariant and variant features, bypassing the difficulty of direct causal discovery from raw data.

We summarize our main contributions as follows:
\begin{itemize}
    \item[1.] 
    We propose FedSDWC, a novel FL model grounded in weak causal inference. It simultaneously addresses OOD generalization and detection by fusing invariant and variant features through a modeled weak causal relationship. Crucially, our approach relaxes the strict invariance assumptions common in traditional methods, enabling more flexible and robust feature utilization.
    \item[2.] 
    We design a novel intervention-based learning strategy to capture the weak causal dependency between invariant and variant features. Theoretically, we provide a tight generalization error bound for FedSDWC. 
    More significantly, we are the first to establish a formal connection between the generalization of causal representations in FL and the clients' prior distributions, addressing a key gap in the literature.
    
    \item[3.] 
    Extensive experiments on multiple benchmark datasets demonstrate that FedSDWC significantly outperforms state-of-the-art (SOTA) models in both OOD generalization and detection, particularly under complex scenarios.
\end{itemize}

\section{Related Works}
\subsection{Federated Learning with Non-IID data}
Data heterogeneity across clients is a primary obstacle in FL, significantly degrading the performance of standard algorithms like FedAvg \citep{mcmahan2017communication}. One line of work aims to create a more robust global model. For instance, FedProx \citep{li2020federated} introduces a regularization term to constrain client model deviation, while FRAug \citep{chen2023fraug} uses representation augmentation with a shared generator to capture cross-client consistency.
A parallel approach is personalized FL, which tailors models to individual clients. Methods like FedL2P \citep{lee2024fedl2p} leverage a meta-network to learn client-specific parameters, and pFedBreD \citep{shi2024prior} decouples personalized priors to improve adaptability. 
While effective for data heterogeneity, these methods lack dedicated mechanisms for OOD data, limiting their generalization performance.

\subsection{OOD Generalization and Detection}
OOD generalization aims to learn robust models by extracting invariant feature-label relationships amidst covariate shifts to ensure reliable deployment \citep{lv2023duet, liao2024foogd, mildner2025federated, nguyen2025federated}. 
Research in this area primarily follows three paradigms. Invariant learning, such as IRM, seeks invariant representations across different training environments \citep{arjovsky2019invariant, guo2023out, li2022learning}. Disentangled learning separates data into stable and variant semantic factors to build robust representations \citep{kong2022partial, bai2024diprompt}, while causal inference leverages tools like structural causal models to eliminate spurious correlations \citep{gui2024joint, zhangcausality, guo2025federated}.
Complementary to this, OOD detection focuses on identifying unknown or novel samples during inference. Prominent approaches include classification-based methods, which utilize model outputs like softmax probabilities \citep{djurisic2023extremely, linderman2023fine, park2023nearest, hendrycks2016baseline}; distance-based methods, which measure a sample's distance to class prototypes, often using the Mahalanobis distance \citep{sun2022out, galesso2023far, ming2024does}; and density-based methods, which model the in-distribution data with techniques like variational autoencoders or flows \citep{wang2022vim, yang2023full, wu2023discriminating}.
Existing methods treat OOD generalization and detection as separate tasks and are constrained by strict invariance assumptions. 
We overcom these limitations with a unified, weak causal framework that leverages both invariant and variant features to improve performance on both tasks simultaneously.

\section{Methodology}
\label{sec:method}
\subsection{Problem Setting}\label{sec:3.1}
\textbf{OOD Generalization and Detection Objective in FL.} 
In real-world FL deployment scenarios, each client \( c \) possesses its own dataset \(\mathcal{D}_c\), leading to two possible cases:
1) The client \( c \) participating in training may have limited data available for model training due to various reasons, such as a large volume of data or the addition of new clients later.
The data used for training is referred to as $\mathcal{D}_c^\text{ID}$. The remaining data may contain covariate-shifted data $\mathcal{D}_c^\text{ID-C}$ and semantic-shifted data \( \mathcal{D}_c^\text{ID-S}\). 
2) For clients not participating in training, their data composition is similar to that of participating clients and may also contain the three aforementioned types of components.
Thus, the dataset of client \( c \) can be represented as \( \mathcal{D}_c = \mathcal{D}_c^\text{ID} + \mathcal{D}_c^\text{ID-C} + \mathcal{D}_c^\text{ID-S} \).
Our objective is:
\begin{equation}
\begin{aligned}
\arg \min_{\theta} \sum_{c=1}^C w_c 
\mathbb{E}_{x \sim p_{\mathcal{D}_c}} \left[\mathcal{L}_c(\theta; \mathcal{D}_c)\right],
\end{aligned}
\end{equation}
where \( w_c \) denotes the weight proportion of the \( c \)-th client, and \( \theta \) represents the model parameters, including those of the classification model and the detector. The \( \mathcal{L}_c(\theta; \mathcal{D}_c) \) can be further decomposed into three components: \( \mathcal{L}_c(\theta; \mathcal{D}_c) = \ell_c^\text{ID} + \ell_c^\text{ID-C} + \ell_c^\text{ID-S} \), where \( \ell_c^\text{ID} \) evaluates the generalization performance of client \( c \) on the \( \mathcal{D}_c^\text{ID} \), \( \ell_c^\text{ID-C} \) evaluates the generalization performance on covariate-shifted data \( \mathcal{D}_c^\text{ID-C} \), and \( \ell_c^\text{ID-S} \) evaluates the detection performance on semantic-shifted data \( \mathcal{D}_c^\text{ID-S} \).
The specific definitions are as follows:
\[
\ell_c^\text{ID} := -\mathbb{E}_{(x, y) \sim p_{\mathcal{D}_c^\text{ID}}} \left[\mathbb{I} \{ y_\text{pred}(f_\theta(x)) = y \} \right],
\]
\[
\ell_c^\text{ID-C} := -\mathbb{E}_{(x, y) \sim p_{\mathcal{D}_c^\text{ID-C}}} \left[\mathbb{I} \{ y_\text{pred}(f_\theta(x)) = y \} \right], 
\]
\[
\ell_c^\text{ID-S} := -\mathbb{E}_{x \sim p_{\mathcal{D}_c^\text{ID-S}}} \left[\mathbb{I} \{ g_\theta(x) = \text{ID} \} \right],
\]
where \( \mathbb{I}(\cdot) \) is the indicator function, \( f_\theta(\cdot) \) is the classification model, and \( g_\theta(\cdot) \) is the detector. 

\begin{figure}
\centering
\includegraphics[width=1.0\columnwidth]{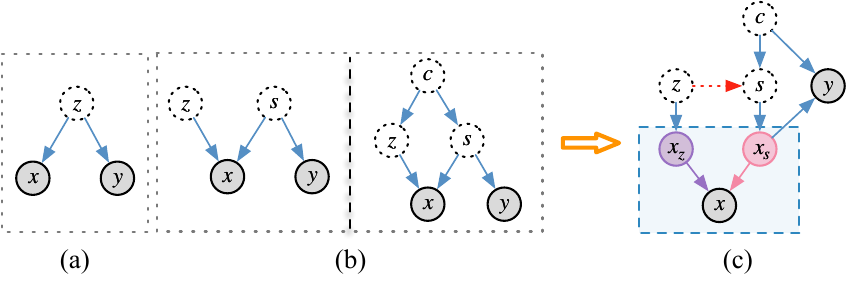}
\caption{Comparison of different causal structures:  
(a) The observed variables \( x \) and \( y \) are solely influenced by the latent variable \( z \), forming a simple causal structure.  
(b) Based on the original structure, an additional latent variable \( s \) is introduced, assuming that \( y \) is solely influenced by \( s \), or assuming that both \( z \) and \( s \) are influenced by a deeper latent variable \( c \).  
(c) \textbf{FedSDWC} improves on the existing models by decomposing \( x \) into invariant features \( x_s \) and environment-related variant features \( x_z \), which are controlled by \( s \) and \( z \), respectively. In inferring \( s \), we consider both the invariant features of \( x_s \) and the weak causal influence of \( z \). Finally, the \( c \) is derived through \( s \), and \( x_s \) is used to infer \( p(y|c, x_s) \).}
\label{fig:causalgraph}
\end{figure}

\subsection{Design of Causal Model in Federated Learning}
Our model design is grounded in a formal definition of causality: \textit{a causal relationship exists between two variables (denoted as ``cause $\rightarrow$ effect") if and only if intervening on the cause (by altering external variables of the system) can potentially change the effect, while the reverse does not hold} \citep{peters2017elements, liu2021learning}. 
Based on this definition of causality, we build our method by step-by-step refining a baseline model, using the image (\(x\)) to label (\(y\)) generation process as an example (Fig. \ref{fig:causalgraph}(c)).

(1) \textbf{Initial Latent Structure.}
In traditional discriminative models, the relationship \(x \to y\) is typically learned directly. However, causal inference focuses more on the latent factors hidden between \(x\) and \(y\). Therefore, we assume that the association between \(x\) and \(y\) mainly originates from a latent variable \(z\) (i.e., a ``pure common cause"). Based on this assumption, the causal graph removes the direct edge \(x \to y\) and retains only the indirect path through \(z\) (Fig. \ref{fig:causalgraph}(a)). 

(2) \textbf{Decomposition of Latent Variables.} 
The initial latent variable \(z\) is further decomposed to separate its causal and non-causal components. It is split into a semantic factor \(s\) (e.g., shape), which is the direct cause of the label \(y\), and a variant factor \(z\) (e.g., background), which captures diversity in the input \(x\). This refinement is represented in the causal graph by removing the direct edge from \(z \to y\), thereby isolating the true causal pathway more precisely  (Fig. \ref{fig:causalgraph}(b)).

(3) \textbf{Eliminating Spurious Correlations.} 
Although \(s\) and \(z\) may exhibit certain statistical correlations in the data (e.g., camels/horses often appear in desert/grassland backgrounds), these are usually spurious correlations. For example, placing a horse in a desert background does not change its label. Therefore, we explicitly distinguish between \(s\) and \(z\) in the causal graph to eliminate their spurious correlations. However, it is unrealistic to completely infer \(s\) and \(z\) from the raw data \(x\), as \(x\) typically contains confounding noise.

To overcome the challenge of inferring latent factors directly from \(x\), the framework first extracts intermediate feature representations. Crucially, unlike methods that discard variant information, this model utilizes both an invariant feature representation \(x_s\) and a variant feature representation \(x_z\) to serve as the basis for inferring \(s\) and \(z\). Recognizing that this feature separation is inevitably imperfect, the model introduces its central innovation: a weak causal relationship (\(z \dashrightarrow s\)) (Fig. \ref{fig:causalgraph}(c)). This edge provides a principled way to model and account for semantic information that may have leaked into the variant features, thereby enhancing the model's causal reasoning and generalization performance.

\subsection{Method for OOD Generalization and Detection }
During training, the model leverages only the ID data, \(\mathcal{D}_c^\text{ID}\), from each participating client \(c \in \mathcal{C}_{par}\).
Through server-side aggregation, all clients fit the global causal graph \(p := \langle p(s|z,c), p(x_s|s), p(x_z|z), p(x|x_s,x_z), p(y|x_s,c)p(c)p(z) \rangle\) by maximizing the likelihood \(\sum_{c\in \mathcal{C}_{par}} \mathbb{E}_{p_c^*(x, y)}[\log p(x, y)]\).
The design of this model is based on the well-known independent causal mechanisms principle. However, in the FL setting, we make certain adaptations to this principle.


\textbf{Federated Learning Causal Invariance}: The proposed causal generative mechanisms \(p(x|c,z)\) and \(p(y|x_s,c)\) remain invariant across all clients and domains, while domain shifts are reflected solely through \(p(z)\).
It is infeasible for each client \(c\) to directly maximize the likelihood \(\mathbb{E}_{p_c^*(x, y)}[\log p(x, y)]\), since $p(x, y) := \int p(c, s, z, x_z, x_s, x, y) \, dcdsdzdx_zdx_s$, where \(p(c, s, z, x_z, x_s, x, y) := p(c)p(z)p(s|c, z) p(x_s|s) p(x_z|z) p(x|x_z, x_s)p(y|x_s, c) \), which is difficult to estimate directly. 
To address this issue, the Evidence Lower Bound (ELBO), defined as \(\mathcal{L}_{p, q_{x_s, x_z, z, s, c|x, y}}(x, y) := \mathbb{E}_{q(x_s, x_z, z, s, c|x, y)}\left[\log \frac{p(x_s, x_z, z, s, c|x, y)}{q(x_s, x_z, z, s, c|x, y)}\right]\), can be introduced as an alternative optimization objective. By introducing an inference model \(q(x_s, x_z, z, s, c|x, y)\), the process of sampling and density estimation can be simplified. Maximizing the ELBO enables \(q(x_s, x_z, z, s, c|x, y)\) to approximate the posterior distribution \(p(x_s, x_z, z, s, c|x, y) := \frac{p(x_s, x_z, z, s, c)}{p(x, y)}\), providing a tighter lower bound for optimizing \(\log p(x, y)\).

However, even after introducing the inference model \(q(x_s, x_z, z, s, c|x, y)\), it remains challenging to directly estimate \(p(y|x)\), making the prediction task difficult. To address this issue, we further introduce the model \(q(x_s, x_z, z, s, c, y|x)\) to approximate the target distribution \(p(x_s, x_z, z, s, c, y|x)\). Through this approximation, it becomes possible to estimate \(y\) by sampling given \(x\).
Specifically, we further transform \(q(x_s, x_z, z, s, c|x, y)\) and express it as:
\(q(x_s, x_z, z, s, c|x, y) = \frac{q(x_s, x_z, z, s, c, y|x)}{q(y|x)}\),
where \(q(y|x) := \int q(x_s, x_z, z, s, c, y|x) \, dcdsdzdx_zdx_s\), which is entirely determined by \(q(x_s, x_z, z, s, c, y|x)\). 
Thus, the ELBO objective for each client \(c\), \(\mathbb{E}_{p_c^*(x)}[\mathcal{L}_p, q_{s, z|x, y}(x, y)]\), can be formulated as:
\begin{equation} \label{eq:elbo1}
\begin{aligned}
   &\mathbb{E}_{p_c^*(x)}\mathbb{E}_{p^*(y|x)} \log q(y|x) + \mathbb{E}_{p^*(x)}\mathbb{E}_{q(c,s,z,x_z,x_s,y|x)} \\
   & \left[ \frac{p^*{(y|x)}}{q(y|x)}\log \frac{p(c,s,z,x_z,x_s,x,y)}{q(c,s,z,x_z,x_s, y|x)} \right].
\end{aligned}
\end{equation}
The first term of the objective function is the negative of the cross-entropy loss, which drives \(q(y|x)\) closer to \(p^*(y|x)\). As this goal is gradually achieved, the second term becomes the ELBO expectation \(\mathbb{E}_{p_c^*(x)}[\mathcal{L}_{p, q_{(x_s, x_z, z, s, c, y|x)}}(x)]\), which works to further approximate \(q_{(x_s, x_z, z, s, c, y|x)}\) to \(p(x_s, x_z, z, s, c, y|x)\), and ensures \(p(x)\) approaches \(p^*(x)\).
Moreover, based on our causal graph (Fig. \ref{fig:causalgraph}(c)), the target distribution can be further decomposed as:
$p(x_s, x_z, z, s, c, y|x) = p(x_s|x)p(x_z|x)p(s|x_s)p(z|x_z, s)p(y|c, x_s)$,
where \(p(y|c, x_s)\) is a known part of the model. Therefore, we can simplify \(q_{(x_s, x_z, z, s, c, y|x)}\) using the inference models \(q(x_s|x)\), \(q(x_z|x)\), \(q(s|x_s)\), and \(q(z|x_z, s)\).
This allows us to further rewrite Equation (\ref{eq:elbo1}) as:
\begin{equation} \label{eq:elbo2}
\begin{aligned}
&\mathbb{E}_{p_c^*(x)}[\mathcal{L}_p, q_{s, z|x, y}(x, y)] = \mathbb{E}_{p^*(x,y)}\log q(y|x)
\\
&+ \mathbb{E}_{p^*(x,y)}\left[\frac{1}{q(y|x)} \mathbb{E}_{q(c,s,z,x_z,x_s|x)} p(y|c,x_s) \right. \\
&\left. \cdot  \log \frac{p(s|z,c)p(z)p(c)p(x_z|z)p(x_s|s)p(x|x_z,x_s)} {q(c,s,z,x_z,x_s|x)} \right].
\end{aligned}
\end{equation}
The above expectations can be estimated using the reparameterization trick combined with the Monte Carlo method. The objective function for client \(c\) is expressed as \(\mathcal{L}^{c}_{elbo}\). 
The detailed derivation can be found in Appendix A.

\subsection{Interventional Learning of Weak Causality}
In our model, the relationship between the latent factors—the variant-derived \(z\) and the invariant-derived \(s\)—is learned through ELBO-based optimization after they are disentangled from the input \(x\). It is important to note that we relax a commonly used assumption in the invariant learning literature \citep{arjovsky2019invariant, guo2023out}, which states that there exists a representation function \(\Phi(\cdot)\) such that for all clients \(c, c' \in \mathcal{C}_{\text{all}}\) and for any \(\hat{z}\) within the intersection of the support sets \(\text{supp}(\mathbb{P}(\Phi(X^c))) \cap \text{supp}(\mathbb{P}(\Phi(X^{c'})))\), the following relationship holds:  
\[
    \mathbb{E}_{X^c, Y^c}[Y^c | \Phi(X^c) = z] = \mathbb{E}_{X^{c'}, Y^{c'}}[Y^{c'} | \Phi(X^{c'}) = z].
\]
We argue that achieving such perfect feature disentanglement is impractical in real-world scenarios. Instead, we posit that the variant (\(x_z\)) and invariant (\(x_s\)) features can only be partially decoupled. This leads to our central hypothesis: a weak causal relationship exists from $z$ to $s$, capturing the subtle but non-negligible influence that arises from this imperfect separation.
The key challenge is to learn this weak causal relationship effectively. To address this, we design a novel, intervention-based objective function aimed at capturing the system's response to small perturbations of the variant factor. We formalize this as the interventional consistency loss, denoted as \(\mathcal{L}_{ic}\). Specifically, for any given sample $x$, we first decouple it into its invariant feature $x_s$ and variant feature $x_z$. 
The intervention involves perturbing the variant feature \(x_z\) with scaled gaussian noise, yielding the intervened feature \(\hat{x}_z = x_z + \alpha \epsilon\), where \(\epsilon \sim \mathcal{N}(0, I)\).
The loss penalizes the divergence between predictions on $(x_s, x_z)$ and $(x_s, \hat{x}_z)$.
We define the loss using KL divergence to measure the discrepancy:
\begin{equation} \label{eq:lic_loss_noise}
\mathcal{L}_{ic} = \mathbb{E}_{x \sim \mathcal{D}} \left[ \text{KL} \left( q(y | x_{s}, x_{z}) \, || \, q(y | x_{s}, \hat{x}_{z}) \right) \right]
\end{equation}
By minimizing \(\mathcal{L}_{ic}\), we compel the model to produce consistent predictions that are robust to \(\alpha\)-scaled perturbations in the variant features. This directly constrains the influence flowing through the \(z \dashrightarrow s\) causal pathway. The scaling factor \(\alpha\) acts as a crucial hyperparameter, controlling the strength of this regularization: a larger \(\alpha\) enforces a stricter invariance, pushing the model to learn a weaker causal link.
\begin{figure}
\centering
\includegraphics[width=1\linewidth]{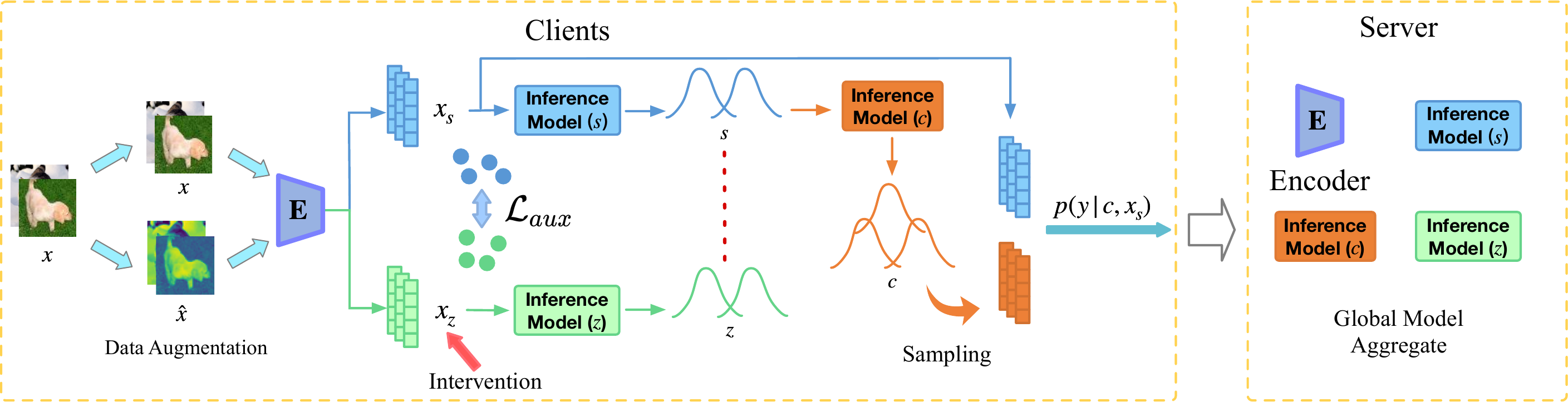}
\caption{Framework of FedSDWC.}
\label{fig:causalmodelframwork}
\end{figure}

\subsection{Model Framework}
The FedSDWC model architecture, depicted in Fig. \ref{fig:causalmodelframwork}.
On the client side, input data $x$ first undergoes Fourier augmentation \citep{xu2021fourier}. This augmented data is then fed into an encoder (WideResNet for CIFAR datasets, ResNet-18 for TinyImageNet), in conjunction with an auxiliary learning method. During training, an intervention strategy is applied to improve generalization. Three MLP-based inference models estimate the distributions of latent factors ($s$, $z$, and $c$) using Gaussian Mixture Models.
Finally, the server aggregates the updated client models using the FedAvg algorithm. The complete training procedure is detailed in Algorithm 1.

\begin{algorithm}[ht]
   \caption{Training procedure of FedSDWC}
   \label{alg:1}
\begin{algorithmic}
   \STATE {\bfseries Input:} Communication rounds \(T\), set of participating clients \(\mathcal{C}_{\text{par}}\), local steps \(E\), batch size \(B\).
   \STATE {\bfseries Ouput: } Optimized global model parameters $\theta_T$.
   \STATE {\bfseries Server Executes. }
   \STATE Initialize global model parameters ($\theta_T$).
   \FOR{$t=1$ {\bfseries to} $T$}
   \FOR{$c \in \mathcal{C}_{\text{par}}$ in parallel do}
   \STATE Send the model \( \theta_t \) to the client \( c \).
   \STATE $\theta_t^c \leftarrow$ Client Executes(c, $\theta_t$).
   \ENDFOR
   \STATE  $ \theta_{t+1}=\sum_{c\in \mathcal{C}_{par}} w_c \theta_t^k$.
   \ENDFOR
   \STATE \textbf{Return} $\theta_T$
   
\STATE {\bfseries Client executes (c, {$\theta_t $}): }
\STATE $\theta_t^c \leftarrow \theta_t$ // Initialize local model with global parameters
\FOR{$e=1$ {\bfseries to} $E$}
\FOR{batch of sample ($X_{1:B}, Y_{1:B} \in \mathcal{D}_c^{\text{ID}}$)}
\STATE$\hat{X}_{1:B}$ = Perform Fourier augmentation on \( X_{1:B} \).
%
\STATE $\mathcal{L}_{total} \leftarrow \mathcal{L}_{elbo}^c(\hat{x}, y; \theta_t^c) + \mathcal{L}_{ic}(\hat{x}, y; \theta_t^c)$ 
\STATE Update $\theta_t^c$ using gradient descent on \(\mathcal{L}_{total}\)
\ENDFOR
\ENDFOR
\STATE \textbf{Return} $\theta_t^c$
\end{algorithmic}
\end{algorithm}

\section{Theoretical Analysis}
\label{sec:theo}
\textbf{Assumption 1 (Additive Noise).}
According to causal graph (Fig. \ref{fig:causalgraph}), the data \( x \) and \( y \) for each client follow the conditional distributions \( p(x | c, z) = p_{\mu}(x - f(c, z)) \) and \( p(y | c, z) = p_{\epsilon}(y - h(c, z)) \), where \( \mu \) and \( \epsilon \) are independent random variables. The function \( f \) is bijective, while \( h \) is injective. For categorical variables \( y \), the conditional distribution can be expressed as \( p(y | c, z) = \text{Cat}(y \mid h(c, z)) \). Furthermore, the nonlinear functions \( f \) and \( h \) have bounded third-order derivatives.

\textbf{Assumption 2.} 
The noise distribution \( p_{\mu} \) for each client has an almost everywhere (a.e.) nonzero characteristic function, such as a Gaussian distribution. 

\textbf{Theorem 1 (OOD generalization error).} 
Under the conditions of Assumptions 1 and 2, the globally aggregated causal model \( p_g \) in FL and the OOD model \( \tilde{p} \) share the same generative mechanism. However, due to environmental differences, the prior distributions \( p_k \) of different clients may vary. For all client data \( x \in \text{supp}(p_{k, x}) \cap \text{supp}(\tilde{p}_x) \), where \( k \in \mathcal{C}_{all} \), the following holds:
\begin{equation}
    \begin{aligned}
        &\mathbb{E}_{\tilde{p}(x)} \left| \mathbb{E}_g \left[ y|x \right] - \tilde{\mathbb{E}} \left[ y|x \right] \right| \\
        &\leq \sigma_{\mu}^2 \mathbb{E}_{\tilde{p}(x)}\left\| \nabla \sum_{k \in \mathcal{C}_{par}}w_k \log \frac{p_k(f^{-1}(x))}{\tilde{p}(f^{-1}(x))} \right\|_2 \\
        &\left\| \mathcal{J}_{f^{-1}(x)} \right\|_2  \|\nabla h \|_2 \big|_{p_k := p_cp_z. (c, z) \sim f^{-1}(x_k)},
    \end{aligned}
\end{equation}
where, \( \mathcal{J}_{f^{-1}}s \) represents the Jacobian determinant of \( f^{-1} \). \( f \) and \( h \) are the mapping functions in the noise addition assumption, where \( f \) satisfies bijectivity, \( h \) satisfies injectivity, and both are three times continuously differentiable.

\textbf{Remark 1:} This bound indicates that when the global causal mechanism \( p(x|z,c) \) is sufficiently strong (i.e., with a smaller \( \sigma_{\mu} \)), it dominates the effect of prior changes, effectively reducing generalization error. 
The term  
\(
\mathbb{E}_{\tilde{p}(x)}\left\| \nabla \sum_{k \in \mathcal{C}_{par}} w_k \log \frac{p_k(f^{-1}(x))}{\tilde{p}(f^{-1}(x))} \right\|_2
\)
can be interpreted as the Fisher divergence, which measures the difference between the prior distributions of each client and the OOD prior. It can also be used to assess the impact of ``OOD-ness" on prediction performance. When the prior distribution of some clients results in a smaller Fisher divergence, the generalization error is correspondingly reduced.  
Furthermore, previous studies have shown that the Fisher divergence is similar in nature to the forward KL divergence and is highly sensitive to insufficiently covered distribution regions \citep{durkan2021maximum}. Consequently, when certain clients have uncovered regions, the term \( \log \left( p_{k}/{\tilde{p}} \right) \) may approach infinity, leading to an increase in generalization error. 
\section{Experiments}
\subsection{Experimental Setups}
\textbf{Datasets.} Following SCONE \citep{bai2023feed} and FOOGD \citep{liao2024foogd}, we select the clear versions of CIFAR-10, CIFAR-100 \citep{krizhevsky2009learning}, and TinyImageNet \citep{le2015tiny}, as ID datasets. To perform OOD generalization, we use corresponding synthetic covariate-shifted datasets as ID-C datasets, which were processed with 15 common image corruption methods. Additionally, we applied 4 extra corruption types to CIFAR-10-C and CIFAR-100-C \citep{hendrycks2018benchmarking}. For OOD detection, we chose five external image datasets: LSUN-Crop, LSUN-Resize \citep{yu2015lsun} , Textures \citep{cimpoi2014describing}, SVHN \citep{netzer2011reading}, and iSUN \citep{xu2015turkergaze} to evaluate the model's performance.  
To evaluate generalization on unseen clients, we use the PACS \citep{li2017deeper} dataset, using one domain as an OOD test domain. Dataset simulation details are in Appendix C.

\textbf{Baseline Methods.} We compared FedSDWC with SOTA FL baselines for OOD detection (FedLN \citep{wei2022mitigating}, FOSTER \citep{yu2023turning}, FedATOL \citep{zheng2023out}) and OOD generalization (FedT3A \citep{iwasawa2021test}, FedIIR \citep{guo2023out}, FedTHE \citep{jiang2022test}, FOOGD \citep{liao2024foogd}, FedICON \citep{tan2023heterogeneity}, PerAda \citep{xie2024perada}, FedCiR \citep{li2024fedcir}). The classical FedAvg \citep{mcmahan2017communication} was also used.

\textbf{Evaluation Metrics.} 
To evaluate the model's generalization on ID and OOD data, we report the accuracy on ID and OOD data, denoted as \textbf{ID-Acc.} and \textbf{ID-C-Acc.}, respectively.
For OOD detection, we use two metrics: the Area Under the ROC Curve (AUROC, higher is better) and the False Positive Rate at a 95\% True Positive Rate (FPR95, lower is better).
Details of the experimental setup are available in Appendix C and the open-source code.

\begin{table*}[ht]
\centering
\setlength{\tabcolsep}{1mm}
\small
\begin{tabular}{cccccccccccc}
\toprule
Corruption & \multicolumn{11}{c}{Method}  \\
\cmidrule{2-12}
 Type & FedAvg & FedLN & FOSTER & FedATOL  & FedT3A & FedIIR & FedTHE & FOOGD & PerAda & FedICON & Ours \\
\midrule
None & 68.03 & 75.24 & 90.22 & 55.93 & 68.03 & 68.26 & 91.05 & 75.09 & 82.92 & 89.06 & \textbf{94.37}  \\
Brightness & 65.44 & 71.77 & 88.70 & 54.44 & 61.52 & 66.12 & 89.71 & 73.71 & 79.00 & 89.18 & \textbf{93.77}\\
Spatter & 62.18 & 67.33 & 85.63 & 51.54 & 55.25 & 60.97 & \textbf{87.66} & 65.31 & 76.16 & 85.23 & 86.79\\
Gaussian Blur & 46.86 & 55.64 & 81.88 & 43.23 & 48.52 & 47.51 & 81.32 & 53.26 & 66.95 & 85.08 & \textbf{86.26}\\
Saturate & 63.62 & 71.76 & 87.06 & 54.39 & 58.41 & 63.32 & 88.73 & 71.98 & 79.73 & 88.64& \textbf{93.29}\\
Speckle Noise & 52.25 & 54.30 & 78.63 & 40.03 & 48.38 & 53.20 & 79.72 & 57.72 & 64.09 & 80.43 & \textbf{83.55}\\
Zoom Blur & 45.15 & 54.88 & 82.41 & 42.33 & 49.48 & 46.57 & 82.07 & 52.97 & 64.80 & 86.05& \textbf{86.87}\\
Fog & 53.89 & 60.82 & 83.35 & 48.17 & 49.52 & 54.85 & 83.35 & 60.96 &  66.09 & 86.35 & \textbf{90.36}\\
Shot Noise & 52.73 & 54.55 & 78.94 & 39.57 & 48.82 & 53.09 & 80.31 & 58.31 & 64.71 & 80.34 & \textbf{84.27} \\
Frosted Glass Blur & 43.13 & 42.33 & 75.03 & 28.97 & 40.41 & 44.53 & \textbf{75.42} & 45.80 & 69.26 & 70.16 & 73.48 \\
Gaussian Noise & 48.66 & 50.25 & 76.80 & 35.20 & 45.27 & 49.15 & 78.37 & 53.92 & 61.09 & 76.82 & \textbf{82.14} \\
Motion Blur & 41.30 & 52.65 & 81.76 & 41.52 & 45.51 & 44.23 & 80.10 & 51.05 & 64.78 & 81.22  & \textbf{87.92}\\
Snow & 54.80 & 60.55 & 83.05 & 45.64 & 51.52 & 55.52 & 83.43 & 61.90 & 71.56 & 82.32 & \textbf{88.59}\\
Elastic Transform & 52.12 & 61.29 & 84.82 & 45.35 & 52.45 & 53.21 & 84.67 & 59.18 & 72.58 & 83.48 & \textbf{88.24} \\
Defouce Blur & 52.37 & 61.08 & 84.44 & 46.36 & 52.60 & 52.72 & 84.35 & 58.66 & 71.58 & 86.63 & \textbf{88.08} \\
Pixelate & 56.88 & 62.00 & 85.71 &46.20 & 53.34 & 59.10 & 84.74 & 64.37 & 79.17 & 85.41 & \textbf{87.89}\\
Contrast & 41.25 & 45.02 & 72.82 & 38.90 & 36.93 & 41.35 & 71.85 & 49.14 & 52.08 & \textbf{87.09} & 81.48\\
Frost & 56.21 & 58.22 & 82.76 & 41.25 & 52.16 & 55.91 & 82.31 & 63.84 & 69.26 & 83.04 & \textbf{88.49} \\
Impulse Noise & 49.32 & 50.52 & 76.52 & 40.36 & 42.79 & 48.45 & \textbf{78.49} & 52.33 & 59.30 & 76.43 & 74.47 \\
Jpeg Compression & 61.56 & 68.61 & 86.68 & 47.94 & 57.64 & 60.46 & 87.50 & 66.55 & 79.39 & 86.37 & \textbf{89.70} \\
\midrule
Avg. & 53.39 & 58.94 & 82.36 & 44.37 & 50.92 & 53.92 & 82.75 &  59.80 & 69.73 & 83.46 & \textbf{86.50}\\
\bottomrule
\end{tabular}
\caption{The comparison results of federated OOD generalization on Cifar-10 ($\alpha=0.1$).
}
\label{tab:gen:cifa10}
\end{table*}

\subsection{Main Results}
\subsubsection{OOD Generalization.}  
We conducted a comprehensive comparison between the FedSDWC and two types of baselines, performing experiments on CIFAR-10, CIFAR-100, and TinyImageNet datasets. To further verify the generalization ability of the model, we introduced different levels of data contamination under a highly heterogeneous data distribution setting (\(\alpha=0.1\)).
The results are presented in Table \ref{tab:gen:cifa10}, Table D.1, and Table D.2 in Appendix D, respectively.   
The results show that classic Non-IID algorithms like FedAvg have poor generalization performance, struggling to handle data contamination, especially with significant distribution shifts. While FedIIR attempts to improve this by learning invariant features, its effectiveness remains limited in highly heterogeneous environments. Although FedIIR demonstrates some improvement in stability, its adaptability to OOD samples is still insufficient.  

\begin{figure}
\centering
\includegraphics[width=0.33\columnwidth]{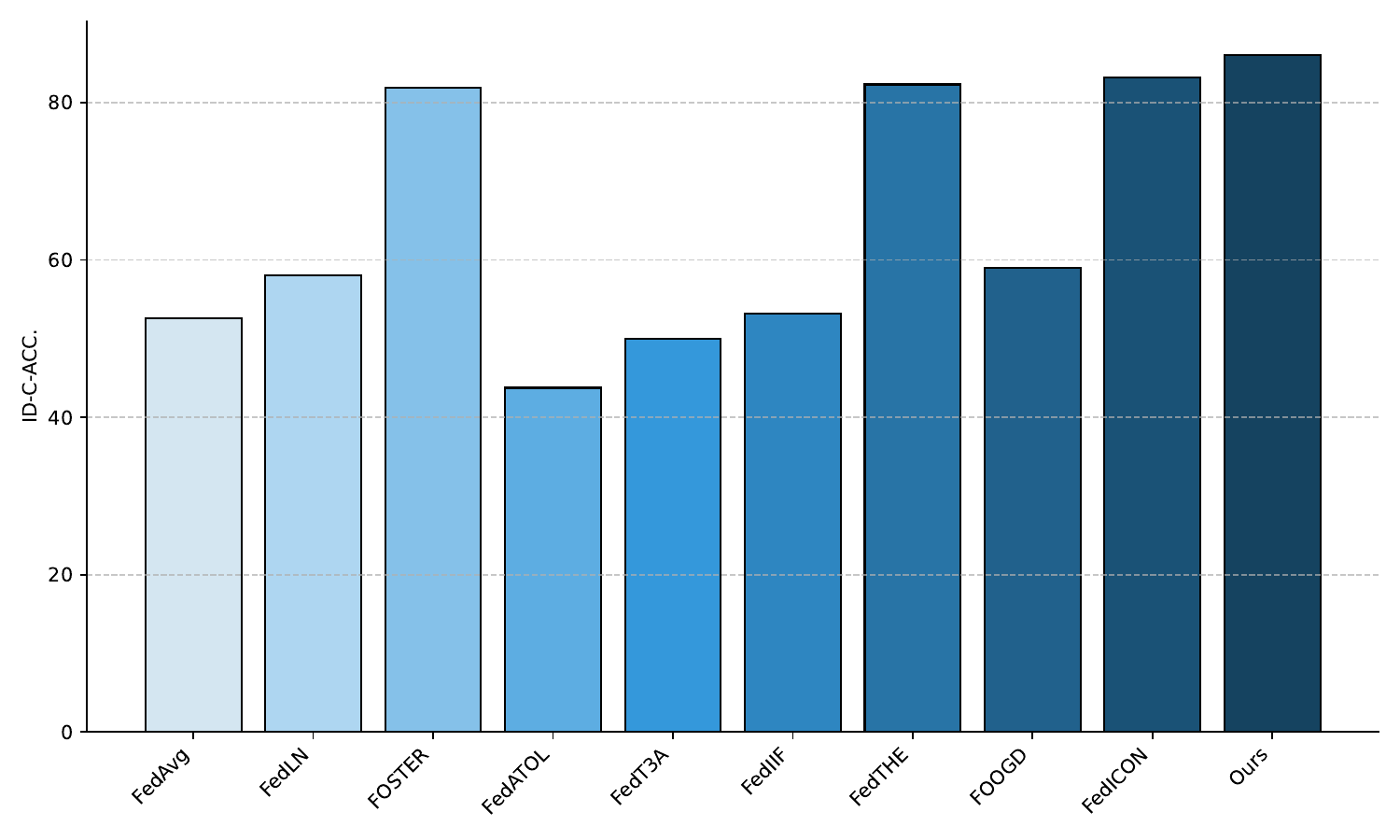} 
\includegraphics[width=0.33\columnwidth]{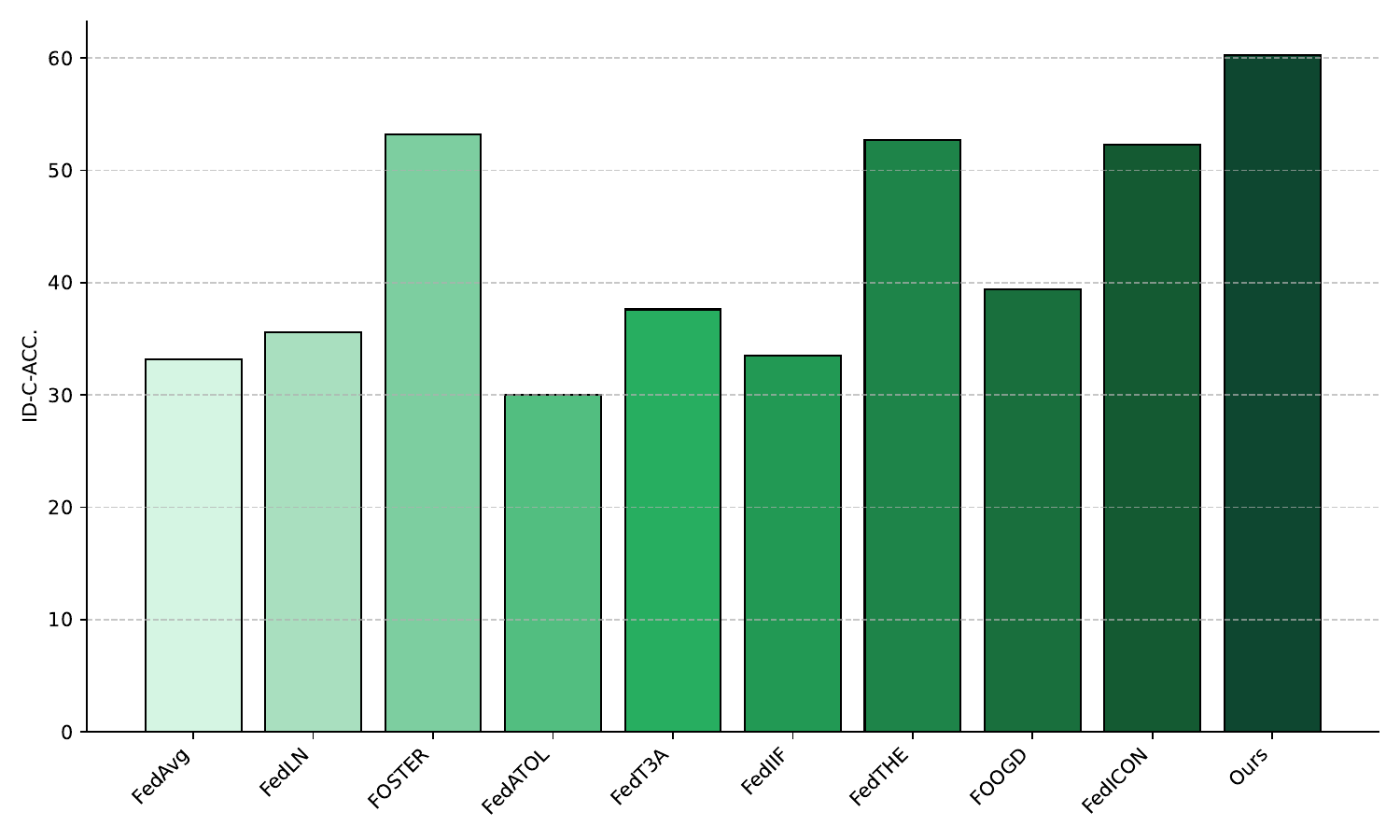}
\caption{Average Generalization Results of the Model on CIFAR-10-C (left) and CIFAR-100-C (right).}
\label{fig:avg-cifar10and100}
\end{figure}

We also found personalized FL models like FedICON and FOSTER to be highly competitive, significantly improving generalization in Non-IID scenarios with data contamination.
This suggests that by learning features tailored to each client's specific distribution, personalized approaches can enhance both global model generalization and local performance.
In contrast, FedSDWC relaxes the strict assumptions of invariant learning by incorporating variant features within a novel causal model designed for Non-IID environments. This approach proved superior across all benchmarks (CIFAR-10, CIFAR-100, and TinyImageNet), where it consistently outperformed existing baselines. FedSDWC demonstrated significantly greater robustness and stability, particularly when handling OOD samples, achieving overall SOTA performance in generalization.
Fig. \ref{fig:avg-cifar10and100} shows the average accuracy of various methods on the corrupted CIFAR-10-C and CIFAR-100-C. The results clearly demonstrate that FedSDWC's generalization performance significantly outperforms key baselines such as FedAvg and FOSTER.
\begin{table*}[!ht]
\small
\setlength{\tabcolsep}{1mm}
\centering
\begin{tabular}{lcccccccccc}
\toprule
\multirow{2}{*}{Method} & \multicolumn{2}{c}{LSUN-Crop} & \multicolumn{2}{c}{LSUN-Resize} & \multicolumn{2}{c}{Textures} & \multicolumn{2}{c}{SVHN} & \multicolumn{2}{c}{iSUN}\\
\cmidrule(r){2-3}
\cmidrule(r){4-5}
\cmidrule(r){6-7}
\cmidrule(r){8-9}
\cmidrule(r){10-11}
& FPR95$\downarrow$ & AUROC $\uparrow$ & FPR95$\downarrow$ & AUROC $\uparrow$ & FPR95$\downarrow$ & AUROC $\uparrow$ & FPR95$\downarrow$ & AUROC $\uparrow$ & FPR95$\downarrow$ & AUROC $\uparrow$\\
\midrule
FedAvg  & 83.41 & 58.05 & 62.01 & 77.02 & 80.53 & 66.23 & 80.02 & 62.14 & 62.10 & 76.29\\
FedLN   & 56.14 & 84.14 & 61.31 & 78.34 & 93.90 & 71.99 & 70.95 & 76.82 &  66.41 & 76.03\\
FOSTER  & 47.40 & 77.43 & 48.09 & 76.24 & 54.23 & 77.62 & 39.55 & 83.07 &  48.73 & 76.29\\
FedATOL & 49.50 & 86.22 & 64.01 & 79.89 & 66.33 & 78.77 & 85.39 & 82.17 &  61.01 & 80.05\\
FedIIR  & 79.48 & 63.31 & 58.44 & 78.69 & 91.72 & 62.32 & 83.68 & 64.04 & 57.86 & 77.98\\
FedTHE  & 58.14 & 82.04 & 42.95 & 83.46 & 53.58 & 82.19 & 39.22 & 85.95 &  43.72 & 83.50\\
FedICON & 48.22 & 81.28 & 49.05 & 83.30 & 51.57 & 80.96 & \textbf{34.94} & 85.56 &  49.98 & 82.95\\
PerAda & 69.35 & 71.84 & 55.77 &73.64 & 76.81 & 68.25 & 47.31 & 78.34 & 56.97 & 76.71 \\
Ours & \textbf{32.61} & \textbf{88.82} & \textbf{36.69} & \textbf{87.70} & \textbf{37.20} & \textbf{87.86} & 35.46 & \textbf{88.08} & \textbf{36.22} & \textbf{87.89}\\
\bottomrule
\end{tabular}
\caption{Experimental results of federated OOD detection on Cifar-10 ($\alpha=0.1$).}
\label{tab:ooddetection}
\end{table*}

\subsubsection{OOD Detection Performance.}
Our method also excels in the OOD detection task under FL. As shown in Table \ref{tab:ooddetection}, when trained on CIFAR-10 and tested against five external datasets (e.g., LSUN, SVHN), our method consistently outperforms competing approaches like FedAvg and FOSTER, particularly on the key metrics of FPR95 and AUROC. For instance, it achieves the lowest FPR95 values (e.g., 32.61 on LSUN-Crop) and the highest AUROC scores (e.g., 88.82 on LSUN-Crop), demonstrating its superior ability to distinguish between ID and OOD samples.

\begin{figure}[ht]
\centering
\includegraphics[width=0.86\columnwidth]{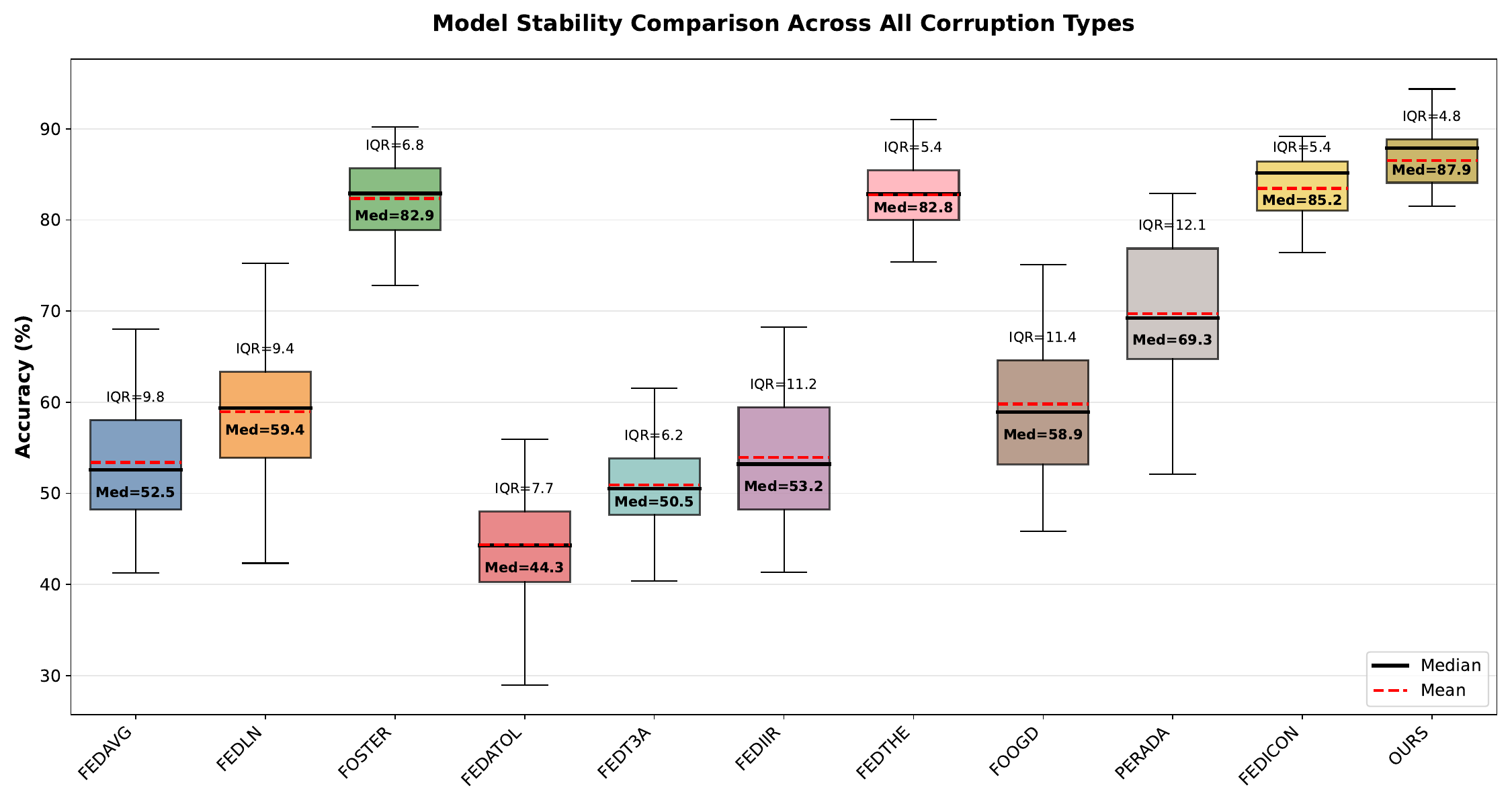}
\caption{Comparison of anti-corruption performance stability across methods on CIFAR-10-C.}
\label{fig:cifar10_box_plot}
\end{figure}

\begin{figure}[ht]
    \centering
    \subfigure[]{
    \begin{minipage}{0.3\columnwidth}
    \centering
    \includegraphics[width=0.5\linewidth]{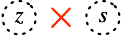}
    \end{minipage}
    }
    \subfigure[]{
    \begin{minipage}{0.3\columnwidth}
    \centering
    \includegraphics[width=0.5\linewidth]{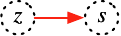}
    \end{minipage}
    }
    \subfigure[]{
    \begin{minipage}{0.3\columnwidth}
    \centering
    \includegraphics[width=0.5\linewidth]{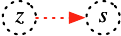}
    \end{minipage}
    }
    \caption{Causal Relationships between $z$ and $s$.}
    \label{fig:zandsrelation}
\end{figure}

\begin{figure}[ht]
\begin{center}
\centerline{
\includegraphics[width=0.4\columnwidth]{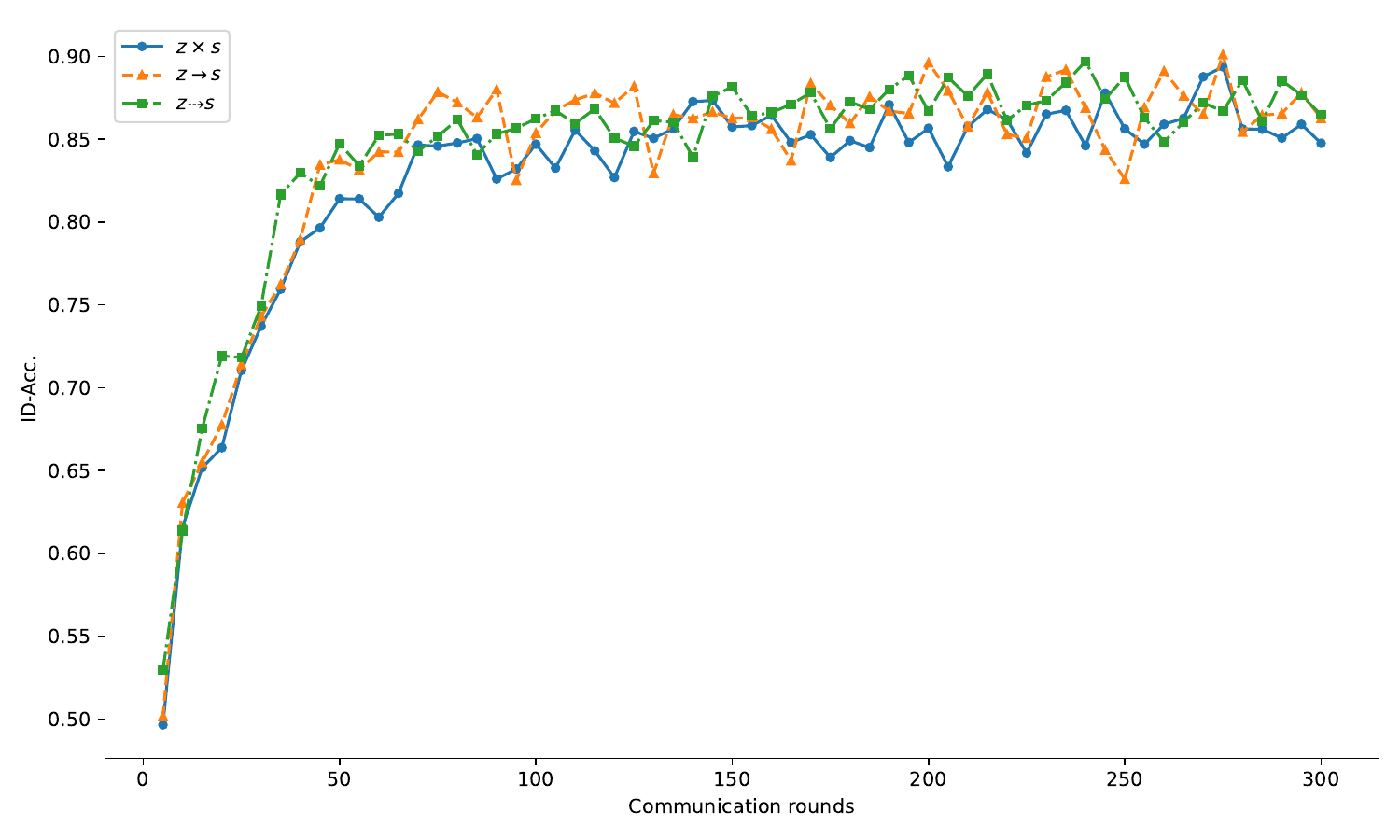}
\includegraphics[width=0.4\columnwidth]{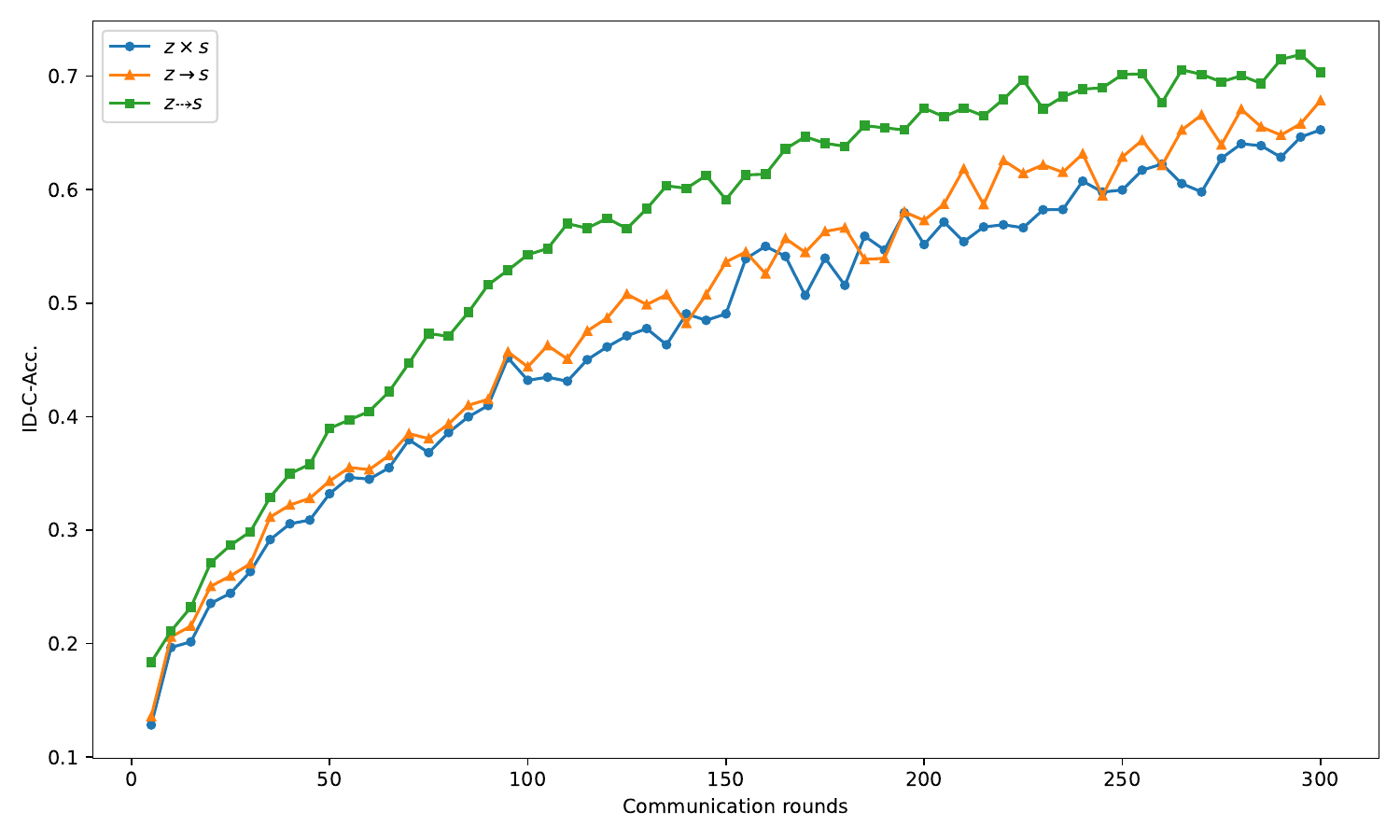}
}
\caption{
Impact of $z-s$ Causality on Generalization.
}
\label{ablation:fig1}
\end{center}
\end{figure}

\begin{table*}[ht]
\centering
\small
\setlength{\tabcolsep}{1mm}
\begin{tabular}{lcccccccccc}
\toprule
\multirow{2}{*}{Method} & \multicolumn{2}{c}{LSUN-Crop} & \multicolumn{2}{c}{LSUN-Resize} & \multicolumn{2}{c}{Textures} & \multicolumn{2}{c}{SVHN} & \multicolumn{2}{c}{iSUN}\\
\cmidrule(r){2-3}
\cmidrule(r){4-5}
\cmidrule(r){6-7}
\cmidrule(r){8-9}
\cmidrule(r){10-11}
& FPR95$\downarrow$ & AUROC $\uparrow$ & FPR95$\downarrow$ & AUROC $\uparrow$ & FPR95$\downarrow$ & AUROC $\uparrow$ & FPR95$\downarrow$ & AUROC $\uparrow$ & FPR95$\downarrow$ & AUROC $\uparrow$\\
\midrule
$z \times s$ & 39.71 & 86.06 & 46.35 & 83.10 & 44.83 & 88.15 & 37.34 & 86.42 & 52.92 & 83.31\\
$z \rightarrow s$ & 37.43 & 86.12 & 43.14 & 83.54 & 34.02 & 88.66 & 36.75 & 87.34 & 48.42 & 83.69 \\
$z \dashrightarrow s $ & \textbf{32.61} & \textbf{88.82} & \textbf{36.69} & \textbf{87.70} & \textbf{37.20} & \textbf{87.86} & \textbf{35.46} & \textbf{88.08} & \textbf{36.22} & \textbf{87.89} \\
\bottomrule
\end{tabular}
\caption{Federated OOD detection Under Different Causal Relationships Between $z$ and $s$ (Cifar-10).}
\label{tab:ablation:tab2}
\end{table*}

\begin{figure}[!ht]
\centering
\subfigure[FedAvg]{
\begin{minipage}{0.22\columnwidth}
    \centering
    \includegraphics[width=1\linewidth]{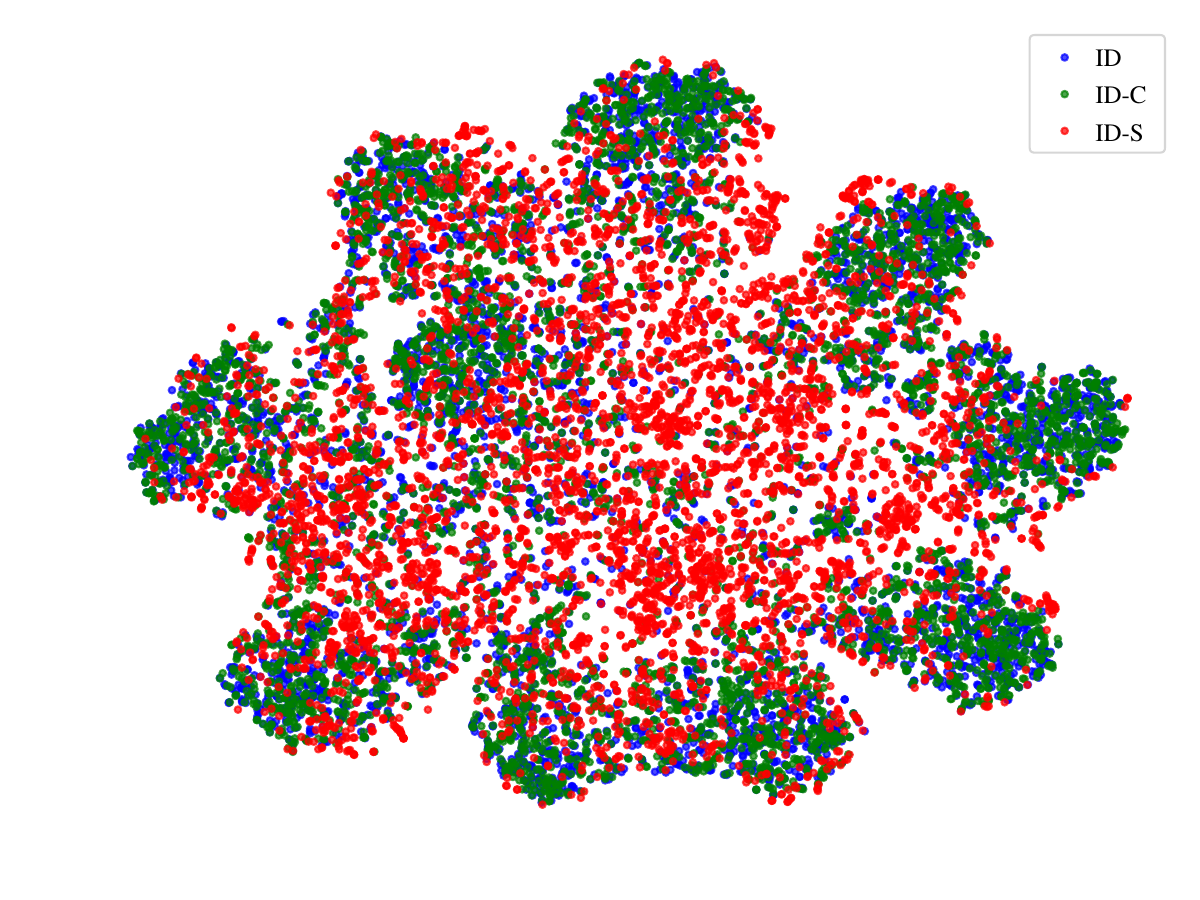}
\end{minipage}
}
\subfigure[PerAda]{\begin{minipage}{0.22\columnwidth}
    \centering
    \includegraphics[width=1\linewidth]{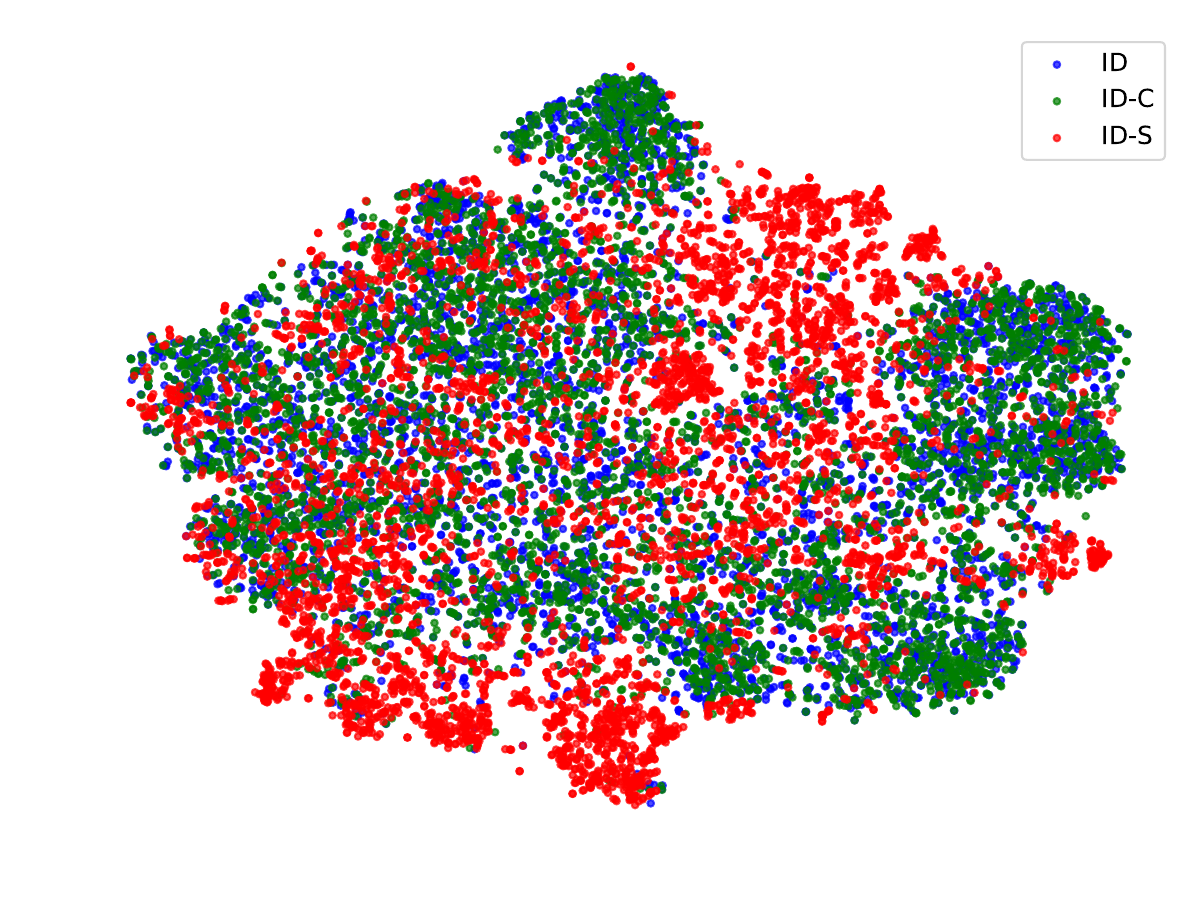}
\end{minipage}}
\subfigure[FedIIR]{
\begin{minipage}{0.22\columnwidth}
    \centering
    \includegraphics[width=1\linewidth]{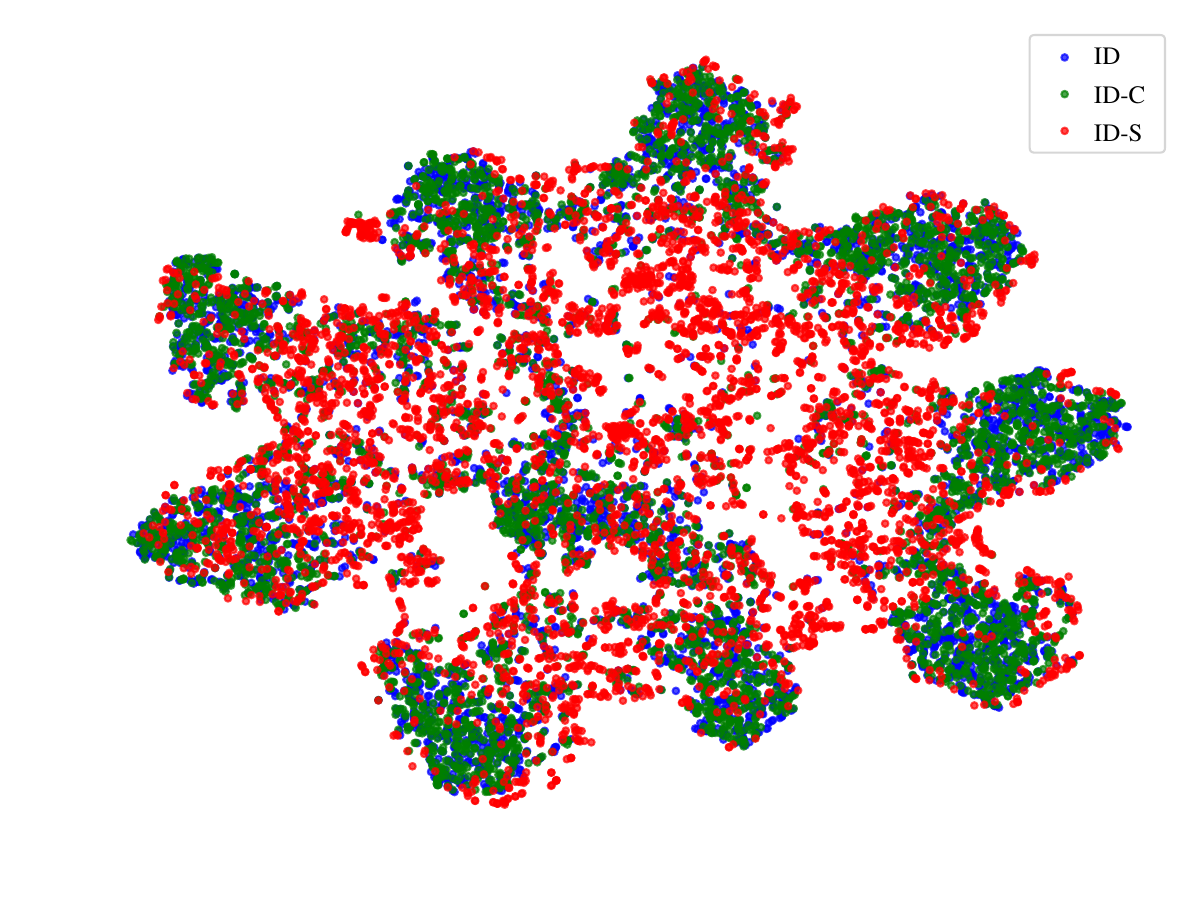}
\end{minipage}
}
\subfigure[Ours]{
\begin{minipage}{0.22\columnwidth}
    \centering
    \includegraphics[width=1\linewidth]{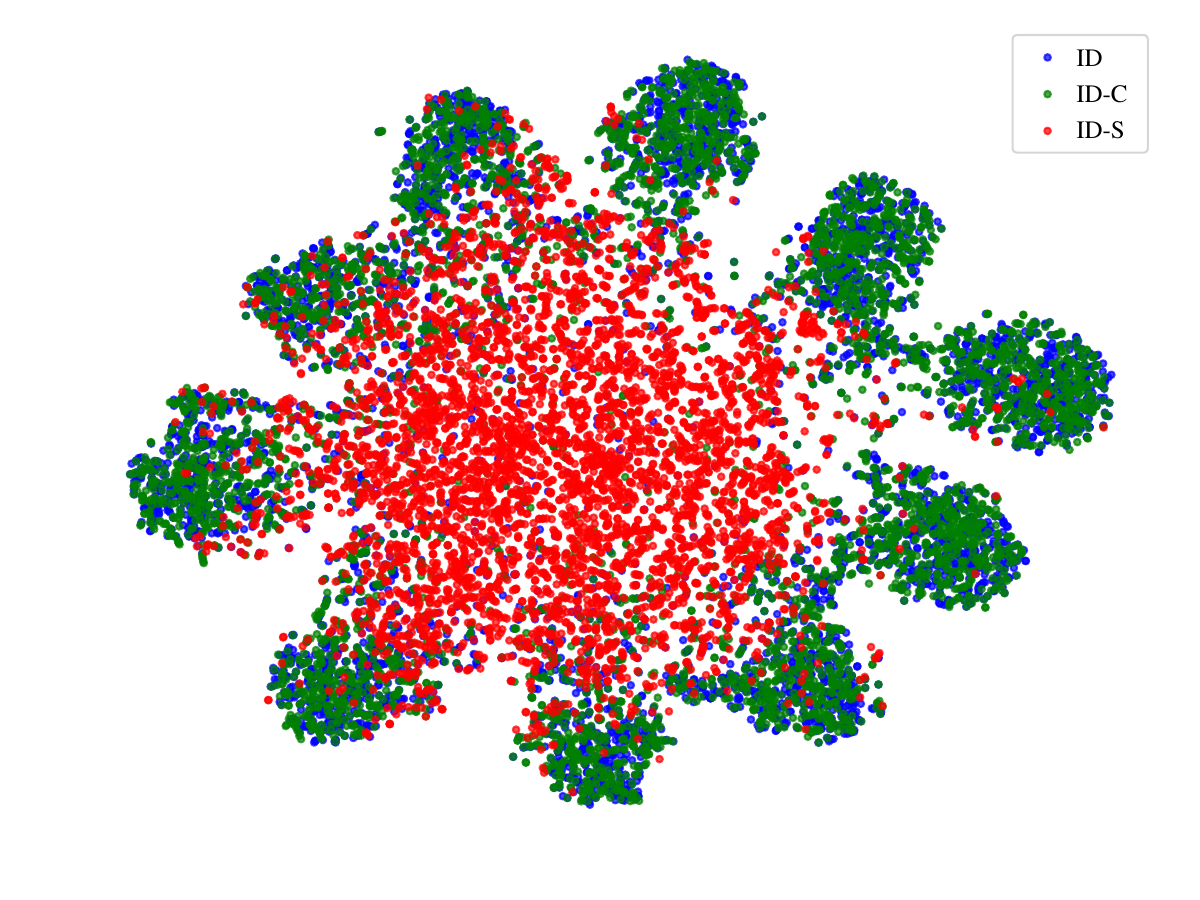}
\end{minipage}
}
\caption{T-SNE visualization of different models ($\alpha=5$).}
\label{fig:t-sne}
\end{figure}

\subsection{Ablation Study}\label{sec:ablation_study}
\textbf{Model Stability Analysis.} We verify our model's robustness through a statistical analysis of its performance distribution on corrupted data (Fig. \ref{fig:cifar10_box_plot}). Our method excels not only in its central tendency, with a median accuracy of 87.9\%, but more critically, in its low dispersion. It achieves an interquartile range (IQR) of 4.8—the smallest among all approaches—indicating that its performance is the most concentrated and consistent across different types of corruption attacks. This demonstrates that our method possesses superior robustness among all evaluated models.

\textbf{Causal Relationship Analysis Between $z$ and $s$.}
We investigate the impact of the causal relationship between $z$ and $s$ on model performance from two perspectives: OOD generalization and detection, as illustrated in Fig. \ref{ablation:fig1} and Table \ref{tab:ablation:tab2}. As shown in Fig. \ref{fig:zandsrelation}, three types of causal relationships exist between $z$ and $s$: no causal relationship ($z \times s$), causal relationship ($z \rightarrow s$), and weak causal relationship ($z \dashrightarrow s$). 
From the left plot in Fig. \ref{ablation:fig1}, it can be observed that regardless of the causal relationship between $z$ and $s$, the model achieves good performance on the ID data. However, the right plot reveals that the causal relationship between $z$ and $s$ significantly affects the OOD generalization of the model. In particular, the weak causal relationship ($z \dashrightarrow s$) exhibits the best performance, with the accuracy steadily increasing throughout the communication rounds. In contrast, the accuracy of the no causal relationship ($z \times s$) and the causal relationship ($z \rightarrow s$) settings is lower and increases at a slower pace.
This phenomenon is further confirmed by Table \ref{tab:ablation:tab2}. 

In the OOD detection task, the model achieves optimal performance only when the causal relationship between $z$ and $s$ is weakened. Specifically, the weak causal relationship ($z \dashrightarrow s$) achieves the best results in key metrics such as FPR95 and AUROC across all datasets. In contrast, the performance under the no causal relationship ($z \times s$) and causal relationship ($z \rightarrow s$) settings is relatively inferior.

\textbf{Visualization.} To explore the characteristics of data distribution in FL OOD generalization/detection models, we present the t-SNE visualization of data representations in Fig. \ref{fig:t-sne}. The experiments, based on the CIFAR-10 ($\alpha=5$), compare FedAvg, PerAda, FedIIR, and our FedSDWC. The results show that our method enables a tighter distribution between ID-C data and ID data, while also establishing clearer decision boundaries between ID data and ID-S data.

\section{Conclusion}
We proposed FedSDWC, a novel FL model that addresses the dual challenges of OOD generalization and detection. Our model uniquely integrates causal inference with invariant learning by modeling a weak causal relationship between variant and invariant features, operationalized through a novel, intervention-based strategy. This enables the model to leverage useful information from variant features while mitigating their spurious correlations. Theoretically, we provide a generalization error bound and establish the first formal link between FL generalization and client data priors. Empirically, FedSDWC achieves SOTA performance, significantly outperforming existing methods.
Future work will focus on enhancing FedSDWC's scalability for massive-scale or extremely heterogeneous data and exploring its application in high-stakes domains such as healthcare.

\input{appendix}

\bibliography{aaai2026}

\end{document}

%% file: appendix.tex
\onecolumn
\appendix
\section*{\centering Appendices}

In the supplementary material, Appendix \ref{appendix:a} provides a detailed derivation of the model's objective, Appendix \ref{appendix:b} presents the proof of the OOD generalization error bound in federated learning, 
Appendix \ref{appendix:c} offers a detailed description of the experimental implementation details, and Appendix \ref{appendix:d} showcases additional experiments.

\setcounter{figure}{0}
\setcounter{equation}{0}
\section{Derivation of FedSDWC Model Objectives}\label{appendix:a}
In federated learning, our goal is to ensure that the global model \( p(x, y) \) matches the data distributions of all participating clients \( p_c^*(x, y) \), where \( c \in \mathcal{C}_{par} \).  
For a given client \( c \), a common and effective approach to achieve this is through maximum likelihood estimation (MLE), allowing the model \( p(x, y) \) to better fit the client's data distribution \( p_c^*(x, y) \). Specifically, this can be formulated as maximizing the following expectation: 
\begin{equation}
\mathbb{E}_{p_c^*(x, y)}[\log p(x, y)].
\end{equation}
The MLE process is equivalent to minimizing the Kullback-Leibler (KL) divergence between the model and the true data distribution:
\begin{equation}
    \text{KL}(p_c^*(x, y) \parallel p(x, y)),
\end{equation}
since \(\mathbb{E}_{p_c^*(x, y)}[\log p_c^*(x, y)]\) is a constant with respect to the model \( p(x, y) \). Thus, this optimization objective ensures that \( p(x, y) \) ultimately converges to \( p_c^*(x, y) \).  

However, the form of the likelihood function:
\begin{equation}
    p(x, y) = \int p(c, s, z, x_s, x_z, x, y) \, dc \, ds \,dz \,dx_s \, dx_z.
\end{equation}
Typically, this involves integrating over high-dimensional latent variables \(c, s, z, x_s \) and \(x_z\), which is intractable in most cases, making the likelihood function difficult to estimate and optimize directly.  
To address this issue, the Variational Expectation-Maximization method has been proposed. This approach simplifies the inference process by introducing a tractable latent variable distribution \(q( c, s, z, x_s, x_z | x, y)\) and constructs a lower bound of the likelihood function, known as the \textit{Evidence Lower BOund} (ELBO).  
The form of this lower bound is given as:

\begin{equation}\label{eq:elbo:4}
\begin{aligned}
\log p(x,y) &= \log \mathbb{E}_{p(c,s,z,x_z,x_s)}\left[ p(c,s,z,x_z,x_s,x,y)\right] \\
&= \log \mathbb{E}_{q(c,s,z,x_z,x_s | x,y)}\left[\frac{p(c,s,z,x_z,x_s,x,y)}{q(c,s,z,x_z,x_s | x,y)}\right] \\
& \geq \mathbb{E}_{q(c,s,z,x_z,x_s | x,y)}\left[\log \frac{p(c,s,z,x_z,x_s,x,y)}{q(c,s,z,x_z,x_s | x,y)}\right].
\end{aligned} 
\end{equation}
Let \(\mathcal{L}_p\) = \(\mathbb{E}_{q(c,s,z,x_z,x_s | x,y)}\left[\log \frac{p(c,s,z,x_z,x_s,x,y)}{q(c,s,z,x_z,x_s | x,y)}\right]\). In variational inference, the function \(\mathcal{L}_p\) and the distribution \(q_{c, s, z, x_z, x_s | x, y}(x, y)\) are referred to as the ELBO. The ELBO serves as a lower bound to estimate the model's log-likelihood function. Its introduction enables efficient inference of the latent variable distribution by optimizing this bound, which in turn allows optimization of the model's parameters. The distribution \(q(c, s, z, x_z, x_s | x, y)\) is referred to as the variational distribution, which is typically instantiated by a separate model (a component of the generative model) known as the inference model.

In the framework of supervised learning in FL, the expected ELBO objective \(\mathbb{E}_{p_c^*(x,y)}[\mathcal{L}_p]\) for each client \(c\) can be regarded as the loss function in traditional supervised learning.
However, as shown in Equation (\ref{eq:elbo:4}), after training, the resulting models include \(p(c,s,z,x_z,x_s,x,y)\) and an approximate model of the posterior distribution \(p(c,s,z,x_z,x_s|x,y)\), denoted as \(q(c,s,z,x_z,x_s|x,y)\). At this point, directly predicting \(p(y|x)\) remains infeasible. To address this issue, we need to reformulate Equation (\ref{eq:elbo:4}) to accommodate the prediction of \(p(y|x)\). Specifically, we introduce a more tractable distribution \(q(c,s,z,x_z,x_s,y|x)\) to model the required variational distribution:
\begin{equation}
   q(c,s,z,x_z,x_s|x,y)=\frac{q(c,s,z,x_z,x_s,y|x)}{q(y|x)}. 
\end{equation}
Here, \(q(y|x)\) is the marginal distribution of \(y\) derived from \(q(c, s, z, x_z, x_s, y|x)\), and its form is:
\begin{equation}
q(y|x) = \int q(c, s, z, x_z, x_s, y|x) \, \mathrm{d}c \, \mathrm{d}s \, \mathrm{d}z \, \mathrm{d}x_z \, \mathrm{d}x_s.
\end{equation}
Through this reformulation, the \(\mathbb{E}_{p_c^*(x,y)}[\mathcal{L}_p]\) objective can be further transformed into:
\begin{equation} \label{eq:elbo:7}
\begin{aligned}
\mathbb{E}_{p_c^*{(x,y)}}\left[\mathcal{L}_p \right] &= \int p_c^*{(x,y)} \frac{q(c,s,z,x_z,x_s,y|x)}{q(y|x)} \\
& \;\;\; \left[\log \frac{p(c,s,z,x_z,x_s,x,y) q(y|x)}{q(c,s,z,x_z,x_s, y|x)}\right]dcdsdzdx_z dx_s dydx \\
& = \int p_c^*{(x,y)} \frac{q(c,s,z,x_z,x_s,y|x)}{q(y|x)}\log q(y|x)dcdsdzdx_z dx_s dydx \\ 
& \;\;\;  + \int p_c^*{(x,y)} \frac{q(c,s,z,x_z,x_s,y|x)}{q(y|x)} \\
& \;\;\;  \left[\log \frac{p(c,s,z,x_z,x_s,x,y)}{q(c,s,z,x_z,x_s, y|x)}\right]dcdsdzdx_z dx_s dydx \\
& = \int p_c^*(x)\left[ \int p^*(y|x) \frac{q(c,s,z,x_z,x_s,y|x)dcdsdzdx_z dx_s}{q(y|x)}\log q(y|x)dy \right]dx \\ 
& \;\;\; + \int p_c^*(x) \left[ \frac{p^*{(y|x)}}{q(y|x)} q(c,s,z,x_z,x_s,y|x) \right. \\
& \left. \;\;\;  \left[\log \frac{p(c,s,z,x_z,x_s,x,y)}{q(c,s,z,x_z,x_s, y|x)}dcdsdzdx_z dx_s \right]  dy\right]dx \\
&=\mathbb{E}_{p_c^*(x)}\mathbb{E}_{p^*(y|x)} \log q(y|x) \\
& \;\;\; + \mathbb{E}_{p_c^*(x)}\mathbb{E}_{q(c,s,z,x_z,x_s,y|x)}\left[ \frac{p^*{(y|x)}}{q(y|x)}\log \frac{p(c,s,z,x_z,x_s,x,y)}{q(c,s,z,x_z,x_s, y|x)} \right]. \\
\end{aligned} 
\end{equation}
Here, \(p^*(y|x) = p_c^*(y|x)\), which is based on the Federated Learning Causal Invariance described in Section 3.3, stating that the proposed causal mechanisms remain invariant across all clients and domains. Based on the causal graph, we can further reformulate Equation (\ref{eq:elbo:7}) as follows:
\begin{equation}\label{app:eq:eq8}
    \begin{aligned}
        \mathbb{E}_{p_c^*{(x,y)}} &= \mathbb{E}_{p_c^*(x,y)}\log q(y|x) \\
        & \;\;\; + \mathbb{E}_{p_c^*(x)}  \int q(c,s,z,x_z,x_s,y|x) \frac{p^*{(y|x)}}{q(y|x)} \\
        & \;\;\; \log \frac{p(c,s,z,x_z,x_s,x,y)}{q(c,s,z,x_z,x_s,y|x)} dcdsdzdx_zdx_sdy \\
        &= \mathbb{E}_{p_c^*(x,y)}\log q(y|x) \\
        &  \;\;\;  + \mathbb{E}_{p_c^*(x)} \int \frac{p^*{(y|x)}}{q(y|x)} \int q(c,s,z,x_z,x_s|x)p(y|c,x_s) \\
        &  \;\;\;  \log \frac{p(c,s,z,x_z,x_s,x)p(y|c,x_s)}{q(c,s,z,x_z,x_s|x)q(y|c,x_s)} dcdsdzdx_zdx_sdy \\
        & = \mathbb{E}_{p_c^*(x,y)}\log q(y|x) \\ 
        &  \;\;\; + \mathbb{E}_{p_c^*(x)} \int \frac{p^*{(y|x)}}{q(y|x)} \int q(c,s,z,x_z,x_s|x)p(y|c,x_s) \\
        &   \;\;\;  \log \frac{p(s|z,c)p(z)p(c)p(x_z|z)p(x_s|s)p(x|x_z,x_s)} {q(c,s,z,x_z,x_s|x)} dcdsdzdx_zdx_sdy \\
        &= \mathbb{E}_{p_c^*(x,y)}\log q(y|x) \\
        &  \;\;\; + \mathbb{E}_{p_c^*(x,y)}\left[\frac{1}{q(y|x)}\mathbb{E}_{q(c,s,z,x_z,x_s|x)}  p(y|c,x_s) \right.\\ 
        & \left.  \;\;\;  \log \frac{p(s|z,c)p(z)p(c)p(x_z|z)p(x_s|s)p(x|x_z,x_s)} {q(c,s,z,x_z,x_s|x)} \right].
    \end{aligned}
\end{equation}
This is the objective function that each client \(c\) needs to optimize on its data distribution \(p_c^*(x, y)\).

\section{Theoretical Analysis: OOD Generalization Error Bound} \label{appendix:b}
In this section, we provide a detailed exposition of Theorem 1 and present a complete proof based on Assumptions 1 and 2. First, we present the assumptions used and restate the Theorem 1, followed by a step-by-step proof to ensure logical consistency and rigor.

\textbf{Assumption 1 (Additive Noise).}
According to causal graph (Fig. 2), the data \( x \) and \( y \) for each client in FL follow the conditional distributions \( p(x | c, z) = p_{\mu}(x - f(c, z)) \) and \( p(y | c, z) = p_{\epsilon}(y - h(c, z)) \), where \( \mu \) and \( \epsilon \) are independent random variables. The function \( f \) is bijective, while \( h \) is injective. For categorical variables \( y \), the conditional distribution can be expressed as \( p(y | c, z) = \text{Cat}(y \mid h(c, z)) \). Furthermore, the nonlinear functions \( f \) and \( h \) have bounded third-order derivatives.

\textbf{Assumption 2} 
The noise distribution \( p_{\mu} \) for each client has an almost everywhere (a.e.) nonzero characteristic function, such as a Gaussian distribution. 

\cite{liu2021learning} proved that under Assumption 2, the causal graph is semantic-identification based on the characteristics of the noise distribution.

\textbf{Theorem 1:} 
Under the conditions of Assumptions 1 and 2, the globally aggregated causal model \( p_g \) in FL and the OOD model \( \tilde{p} \) share the same generative mechanism. However, due to environmental differences, the prior distributions \( p_k \) of different clients may vary. For all client data \( x \in \text{supp}(p_{k, x}) \cap \text{supp}(\tilde{p}_x) \), where \( k \in \mathcal{C}_{all} \), the following holds:
\begin{equation}
    \begin{aligned}
        \mathbb{E}_{\tilde{p}(x)} \left| \mathbb{E}_g \left[ y|x \right] - \tilde{\mathbb{E}} \left[ y|x \right] \right| \leq &  \sigma_{\mu}^2 \mathbb{E}_{\tilde{p}(x)}\left\| \nabla \sum_{k \in \mathcal{C}_{par}}w_k \log \frac{p_k(f^{-1}(x))}{\tilde{p}(f^{-1}(x))} \right\|_2 \\
        & \left\| \mathcal{J}_{f^{-1}(x)} \right\|_2  \|\nabla h \|_2 \big|_{p_k := p_cp_z. (c, z) \sim f^{-1}(x_k)},
    \end{aligned}
\end{equation}
where, \( \mathcal{J}_{f^{-1}}s \) represents the Jacobian determinant of \( f^{-1} \). \( f \) and \( h \) are the mapping functions in the noise addition assumption, where \( f \) satisfies bijectivity, \( h \) satisfies injectivity, and both are three times continuously differentiable.


From the causal graph, it can be observed that in each client, the prior variables \( c \) and \( z \) influence the data \( x \) through the mediator variable \( s \). The joint distribution can be expressed as:
\begin{equation}
 p(x) = \int p(c)p(z)p(s | c, z)p(x_s | s)p(x_z | z)p(x \mid x_s, x_z) \, dc \, dz \, ds \, dx_s \, dx_z.
\end{equation}

It follows that the data \( x \) is primarily determined by the prior distributions of \( c \) and \( z \). Therefore, to simplify the proof process, the above distribution can be further approximated as:
\begin{equation}
p(x) = \int p(x | c, z)p(c)p(z)\;dcdz.
\end{equation}
Based on the additive noise assumption in Assumption 3, 
we further introduce the variable \( v \), which represents the joint variable \( (c, z) \), thereby rewriting the data distribution as:
\begin{equation}
p(x) = \int p(v)p(x | v) \, dv.
\end{equation}
Through the above transformation, we simplify the complex causal relationships into a conditional distribution based on \( v \). According to the additive noise assumption in Assumption 3, the conditional distribution can be expressed as:
\begin{equation}
    p(x | v) = p_{\mu} (x - f(v)).
\end{equation}
Furthermore, we define:
\(\Phi_x(v) := x - f(v)\) and \( \mu = \Phi_x(v)\),
where the function \( \Phi_x(v) \) is invertible, and its inverse mapping is given by:
\[
\Phi_x^{-1}(\mu) = f^{-1}(x - \mu).
\]

Next, we compute the conditional expectation \( \mathbb{E}[y \mid x] \) under the data distribution \( p(x, y) \) for each client \( c \), $c\in \mathcal{C}_{par}$.
\begin{equation}
\begin{aligned}\label{app:eq:13}
\mathbb{E} \left[ y|x \right]
& = \frac{1}{p(x)}\int y p(x,y) \, dy \\
& = \frac{1}{p(x)} \int \int y p(v)p(x|v)p(y|v) dvdy\\
& = \frac{1}{p(x)} \underbrace{\int p(v)p(x|v)\mathbb{E}\left[ y|v \right] dv}_{A}.
\\
\end{aligned}
\end{equation}
For \( p(x) \), we have:
\begin{equation}
    \begin{aligned}
p(x) 
& = \int p(v)p(x | v) \, dv = \int p(v)p_{\mu} (x - f(v)) dv\\
& = \int p(v)p_{\mu} (\Phi_x(v))dv \\
& = \int p(\Phi_x^{-1}(\mu)) p_{\mu}  | J_{\Phi_x^{-1}(\mu)}| d \mu \\
& = \int p(f^{-1}(x - \mu)) p_{\mu} | J_{f^{-1}(x - \mu)}| d\mu.
    \end{aligned}
\end{equation}

Let $p\circ f:= p(f^{-1}(x))$, $\mathcal{J} := | J_{f^{-1}}|$, Then,
\begin{equation}    
\begin{aligned}\label{app:eq:px}
p(x)
& = \int \mathcal{J} p\circ f(x-\mu) p_{\mu} d\mu \\
& = \mathbb{E}_{p_{\mu}} \left[ \mathcal{J} p\circ f -\nabla (\mathcal{J} p\circ f)^T \mu + \frac{1}{2} \mu^T \nabla \nabla (\mathcal{J} p\circ f)\mu + O(\mathbb{E}\left\| \mu \right\|_2^3)\right] \\
& = \mathcal{J} p\circ f + \frac{1}{2}\mathbb{E}_{p_{\mu}}\left[ \mu^T \nabla \nabla (\mathcal{J}p\circ f)\mu 
 \right] +  O(\sigma_\mu^3).
    \end{aligned}
\end{equation}
Using the expansion \( \frac{1}{x+\xi} = \frac{1}{x} - \frac{\xi}{x^2} + O(\xi^2) \), we obtain from Equation (\ref{app:eq:px}):
\begin{equation}\label{app:eq:16}
    \begin{aligned}
        \frac{1}{p(x)} = \frac{1}{\mathcal{J}p\circ f} - \frac{\mathbb{E}_{p_{\mu}}\left[ \mu^T \nabla \nabla (\mathcal{J}p\circ f)\mu 
 \right]}{2(\mathcal{J}p\circ f)^2} +  O(\sigma_\mu^3).
    \end{aligned}
\end{equation}
 
Next, we simplify \( A \) in Equation (\ref{app:eq:13}):
\begin{equation}\label{app:eq:17}
\begin{aligned}
A & = \int p(v)p(x|v)\mathbb{E}\left[ y|v \right] dv \\
& = \int p(v)p_{\mu}(x-f(v))\mathbb{E}\left[ y|v \right]dv \\
& = \int p(\Phi_x^{-1}(\mu))p_{\mu}(\Phi_x(v))\mathbb{E}\left[ y|\Phi_x^{-1}(\mu) \right]dv \\
& = \int p(\Phi_x^{-1}(\mu))p_{\mu} \mathbb{E}\left[ y|\Phi_x^{-1}(\mu) \right] | J_{f^{-1}(x - \mu)}| d\mu \\
& = \int p(f^{-1}(x-\mu)) \mathbb{E}\left[ y|\Phi_x^{-1}(\mu) \right] p_{\mu} | J_{f^{-1}(x - \mu)}| d\mu\\
&= \mathbb{E}_{p_{\mu}} \left[
\mathcal{J}p\circ f(x-\mu) \mathbb{E}\left[ y|\Phi_x^{-1}(\mu) \right] 
\right] \\
& = \mathbb{E}_{p_{\mu}} \left[ \mathcal{J} p\circ f \mathbb{E}[ y|\Phi_x^{-1}(\mu)] -\nabla (\mathcal{J} p\circ f\mathbb{E}[ y|\Phi_x^{-1}(\mu)])^T \mu \right. \\
& \left. \;\;\; + \frac{1}{2} \mu^T \nabla \nabla (\mathcal{J} p\circ f\mathbb{E}[ y|\Phi_x^{-1}(\mu)])\mu + O( \sigma_{\mu}^3)\right] \\
& = \mathbb{E}_{p_{\mu}} \left[ \mathcal{J} p\circ f \mathbb{E}[ y|\Phi_x^{-1}(\mu)] + \frac{1}{2} \mu^T \nabla \nabla (\mathcal{J} p\circ f\mathbb{E}[ y|\Phi_x^{-1}(\mu)])\mu + O( \sigma_{\mu}^3)\right].
\end{aligned}
\end{equation}

Applying Equations (\ref{app:eq:16}, \ref{app:eq:17}) to Equation (\ref{app:eq:13}), we have:
\begin{equation}
\begin{aligned}
\mathbb{E} \left[ y|x \right] & = \frac{A}{p(x)}\\ 
&= \left[ \frac{1}{\mathcal{J}p\circ f} - \frac{\mathbb{E}_{p_{\mu}}\left[ \mu^T \nabla \nabla (\mathcal{J}p\circ f)\mu 
 \right]}{2(\mathcal{J}p\circ f)^2} \right]A \\
 & = \mathbb{E}_{p_{\mu}} \left [
\mathbb{E}[ y|\Phi_x^{-1}(\mu)] + \frac{1}{2}\mu^T 
\nabla \nabla (\mathcal{J} p\circ f\mathbb{E}[ y|\Phi_x^{-1}(\mu)])\mu
 \right ] \\
 & = \mathbb{E}_{p_{\mu}} \left [
\mathbb{E}[ y|\Phi_x^{-1}(\mu)] + \frac{1}{2}\mu^T 
(\nabla \log p\circ f \mathcal{J}\nabla (\mathbb{E}[ y|\Phi_x^{-1}(\mu))^T] \right. \\
& \left. \;\;\; + \nabla \mathbb{E}[ y|\Phi_x^{-1}(\mu)](\nabla \log p\circ f \mathcal{J})^T + \nabla \nabla^Ts \mathbb{E}[ y|\Phi_x^{-1}(\mu)]  )\mu \right ] \\
& \;\;\;+ O(\sigma_{\mu^3}).
\end{aligned}
\end{equation}
According to causal invariance, any two causal models share the same conditional expectation \(\mathbb{E}[ y|\Phi_x^{-1}(\mu) ]\) and \(\mathcal{J}\), since they have the same distributions \( p(x|c,z) \) and \( p(y|c,z) \), thereby sharing the same functions \( f \) and \( h \). Therefore, for any \( x \) belonging to \( \operatorname{supp}(p_{x\in D^{\text{ID}}}) \cap \operatorname{supp}(\tilde{p}_{x\in D^{\text{ID-C}}}) \), we have:

\begin{equation}\label{app:eq:19}
\begin{aligned}
\mathbb{E} \left[ y|x \right] = & \mathbb{E}[ y|\Phi_x^{-1}(\mu)] + \frac{1}{2}\mathbb{E}_{p_{\mu}}  [\mu^T 
(\nabla \log p\circ f \mathcal{J}\nabla (\mathbb{E}[ y|\Phi_x^{-1}(\mu)])^T \\ 
&+ \nabla \mathbb{E}[ y|\Phi_x^{-1}(\mu)](\nabla \log p\circ f \mathcal{J})^T + \nabla \nabla^T \mathbb{E}[ y|\Phi_x^{-1}(\mu)]  )\mu ] \\
& + O(\sigma_{\mu^3}).
\end{aligned}
\end{equation}

\begin{equation}\label{app:eq:20}
\begin{aligned}
\tilde{\mathbb{E}} \left[ y|x \right] = & \mathbb{E}[ y|\Phi_x^{-1}(\mu)] + \frac{1}{2}\mathbb{E}_{p_{\mu}}  [\mu^T 
(\nabla \log \tilde{p}\circ f \mathcal{J}\nabla (\mathbb{E}[ y|\Phi_x^{-1}(\mu)])^T \\ 
&+ \nabla \mathbb{E}[ y|\Phi_x^{-1}(\mu)](\nabla \log \tilde{p}\circ f \mathcal{J})^T + \nabla \nabla^T \mathbb{E}[ y|\Phi_x^{-1}(\mu)]  )\mu ] \\
& + O(\sigma_{\mu^3}). 
\end{aligned}
\end{equation}

In FL, to obtain the global conditional expectation \( \mathbb{E}_g[y \mid x] \), it is necessary to aggregate the local conditional expectations \( \mathbb{E}_k[y \mid x] \;(c \in \mathcal{C}_{par}) \) from all participating clients. According to Equation (\ref{app:eq:19}), the conditional expectation for each client $k$ is given by:

\begin{equation}
\begin{aligned}
\mathbb{E}_k \left[ y|x \right] = & \mathbb{E}[ y|\Phi_x^{-1}(\mu)] + \frac{1}{2}\mathbb{E}_{p_{\mu}}  [\mu^T 
(\nabla \log p_k\circ f \mathcal{J}\nabla (\mathbb{E}[ y|\Phi_x^{-1}(\mu)])^T \\ 
& + \nabla \mathbb{E}[ y|\Phi_x^{-1}(\mu)](\nabla \log p_k\circ f \mathcal{J})^T + \nabla \nabla^T \mathbb{E}[ y|\Phi_x^{-1}(\mu)]  )\mu ] \\
& + O(\sigma_{\mu^3}).
\end{aligned}
\end{equation}

The global conditional expectation \( \mathbb{E}_g[y \mid x] \) is computed by performing a weighted summation of the local conditional expectations from all participating clients in the set \( \mathcal{C}_{par} \), where the sum of the weights satisfies \( \sum_{k \in \mathcal{C}_{par}} w_c = 1 \). The expression is as follows:
\begin{equation}\label{app:eq:22}
    \begin{aligned}
        \mathbb{E}_g[y \mid x] &= \sum_{k \in \mathcal{C}_{par}} w_k \mathbb{E}_k \left[ y \mid x \right] \\
        &=\sum_{k \in \mathcal{C}_{par}} w_k\mathbb{E}[ y|\Phi_x^{-1}(\mu)] 
+ \frac{1}{2}\mathbb{E}_{p_{\mu}}  \left[\mu^T  
\left[\sum_{k \in \mathcal{C}_{par}}w_k\nabla \log p_k\circ f \mathcal{J}\nabla (\mathbb{E}[ y|\Phi_x^{-1}(\mu)])^T \right.\right. \\
& \left.\left. \;\;\; + \sum_{k \in \mathcal{C}_{par}} w_k \nabla \mathbb{E}[ y|\Phi_x^{-1}(\mu)](\nabla \log p_k\circ f \mathcal{J})^T + \sum_{k \in \mathcal{C}_{par}}w_k\nabla \nabla^T \mathbb{E}[ y|\Phi_x^{-1}(\mu)]  \right]\mu \right] \\
& \;\;\; + O(\sigma_{\mu^3}) \\
&= \mathbb{E}[ y|\Phi_x^{-1}(\mu)] + \frac{1}{2} \mathbb{E}_{p_{\mu}} 
\left[\mu^T\left[
\left(\sum_{k \in \mathcal{C}_{par}}w_k\nabla \log p_k\circ f \mathcal{J} \right)\nabla (\mathbb{E}[ y|\Phi_x^{-1}(\mu)])^T \right.\right. \\
& \left.\left. \;\;\; + \nabla \mathbb{E}[ y|\Phi_x^{-1}(\mu)]\sum_{k \in \mathcal{C}_{par}} w_k (\nabla \log p_k\circ f \mathcal{J})^T + \nabla \nabla^T \mathbb{E}[ y|\Phi_x^{-1}(\mu)] 
\right] \mu 
\right]\\
&\;\;\; + O(\sigma_{\mu^3}). 
\end{aligned}
\end{equation}

After obtaining the global conditional expectation \( \mathbb{E}_g[y | x] \), subtracting Equation (\ref{app:eq:22}) from Equation (\ref{app:eq:19}) yields:
\begin{equation}
    \begin{aligned}
       \left| \mathbb{E}_g \left[ y|x \right] - \tilde{\mathbb{E}} \left[ y|x \right] \right| &= \frac{1}{2} \left| \mathbb{E}_{p_{\mu}} \left[
       \mu^T \left(
       \nabla \sum_{k \in \mathcal{C}_{par}}w_k \log \frac{ p_k\circ f}{\tilde{p}\circ f} \left(\nabla \mathbb{E}[ y|\Phi_x^{-1}(\mu)]\right)^T \right. \right. \right. \\
       & \;\;\;\left. \left. \left. + \nabla \mathbb{E}[ y|\Phi_x^{-1}(\mu)] \left(\nabla \sum_{k \in \mathcal{C}_{par}}w_k\log \frac{ p_k\circ f}{ \tilde{p}\circ f}\right)^T 
       \right) \mu
      \right ] \right| \\
      & \leq \frac{1}{2} \mathbb{E}_{p_{\mu}} \left[ \left|
       \mu^T \left(
       \nabla \sum_{k \in \mathcal{C}_{par}}w_k \log \frac{p_k\circ f}{\tilde{p}\circ f} \left(\nabla \mathbb{E}[ y|\Phi_x^{-1}(\mu)]\right)^T \right. \right. \right.\\
       & \;\;\;\left. \left. \left. + \nabla \mathbb{E}[ y|\Phi_x^{-1}(\mu)] \left(\nabla \sum_{k \in \mathcal{C}_{par}}w_k \log \frac{ p_k\circ f}{ \tilde{p}\circ f}\right)^T 
       \right) \mu \right|
      \right ] \\
      & \leq \frac{1}{2} \mathbb{E}_{p_{\mu}} \left[ 
       \|\mu\|^2_2 \left(
       \left\| \nabla \sum_{k \in \mathcal{C}_{par}}w_k \log \frac{p_k\circ f}{\tilde{p}\circ f} \left(\nabla \mathbb{E}[ y|\Phi_x^{-1}(\mu)]\right)^T \right\|_2 \right. \right. \\
       & \left. \left. \;\;\; + \left\| \nabla \mathbb{E}[ y|\Phi_x^{-1}(\mu)] \left(\nabla \sum_{k \in \mathcal{C}_{par}}w_k \log \frac{p_k\circ f}{\tilde{p}\circ f}\right)^T \right\|
       \right) 
      \right ]\\
       & = \underbrace{\left| \nabla \sum_{k \in \mathcal{C}_{par}}w_k \log \frac{p_k\circ f}{ \tilde{p}\circ f} \left(\nabla \mathbb{E}[ y|\Phi_x^{-1}(\mu)]\right)^T \right|} _{B}\mathbb{E}[\mu^T\mu].
    \end{aligned}  
\end{equation}
The term \( B \) can be further computed, which gives:
\begin{equation}
    \begin{aligned}
        B &= \left| \nabla \sum_{k \in \mathcal{C}_{par}}w_k \log \frac{p_k\circ f}{\tilde{p}\circ f} \left(\nabla \mathbb{E}[ y|\Phi_x^{-1}(\mu)]\right)^T \right| \\
        &= \left| \nabla \sum_{k \in \mathcal{C}_{par}}w_k \log \frac{p_k\circ f}{\tilde{p}\circ f} \left(\nabla h(f^{-1}(x-\mu))\right)^T \right| \\
        &= \left| \nabla \sum_{k \in \mathcal{C}_{par}}w_k \log \frac{p_k\circ f}{\tilde{p}\circ f} \mathcal{J}^T \nabla h \mathcal{J}\right| \\
        & \leq \left\| \nabla \sum_{k \in \mathcal{C}_{par}}w_k \log \frac{p_k\circ f}{\tilde{p}\circ f} \right\|_2 \left\| \mathcal{J} \right\|_2  \|\nabla h \|_2 .
    \end{aligned}
\end{equation}
Then,
\begin{equation}
    \begin{aligned}
        \left| \mathbb{E}_g \left[ y|x \right] - \tilde{\mathbb{E}} \left[ y|x \right] \right| \leq \sigma_{\mu}^2\left\| \nabla \sum_{k \in \mathcal{C}_{par}}w_k  \log \frac{p_k\circ f}{\tilde{p}\circ f} \right\|_2 \left\| \mathcal{J} \right\|_2  \|\nabla h \|_2 .
    \end{aligned}
\end{equation}
Here, \( \mathcal{J} = \mathcal{J}_{f^{-1}} \) represents the Jacobian determinant of \( f^{-1} \). Then,   
\begin{equation}
    \begin{aligned}
        \left| \mathbb{E}_g \left[ y|x \right] - \tilde{\mathbb{E}} \left[ y|x \right] \right| &\leq \sigma_{\mu}^2\left\| \nabla \sum_{k \in \mathcal{C}_{par}}w_k \log \frac{p_k(f^{-1}(x))}{ \tilde{p}(f^{-1}(x))} \right\|_2 \left\| \mathcal{J}_{f^{-1}(x)} \right\|_2  \\
        & \|\nabla h \|_2 \big|_{p_k := p_cp_z. (c, z) \sim f^{-1}(x_k)} .
    \end{aligned}
\end{equation}
\setcounter{table}{0}   
\setcounter{figure}{0}
\setcounter{equation}{0}
\counterwithin{figure}{section}
\numberwithin{figure}{section}
\counterwithin{table}{section}
\numberwithin{table}{section}
\section{Experimental Implementation Details}\label{appendix:c}

\subsection{Experimental setup details}    
\textbf{Datasets.} Following SCONE \citep{bai2023feed} and FOOGD \citep{liao2024foogd}, we select the clear versions of CIFAR-10, CIFAR-100 \citep{krizhevsky2009learning}, and TinyImageNet \citep{le2015tiny}, as ID datasets. To perform OOD generalization, we use corresponding synthetic covariate-shifted datasets as ID-C datasets, which were processed with 15 common image corruption methods. Additionally, we applied 4 extra corruption types to CIFAR-10-C and CIFAR-100-C \citep{hendrycks2018benchmarking}. For OOD detection, we chose five external image datasets: LSUN-Crop, LSUN-Resize \citep{yu2015lsun} , Textures \citep{cimpoi2014describing}, SVHN \citep{netzer2011reading}, and iSUN \citep{xu2015turkergaze} to evaluate the model's performance.  
To further evaluate the generalization ability on unseen client data, we also employ the PACS\citep{li2017deeper} dataset, using one domain as an OOD test domain. 

\textbf{Baseline Methods.} To comprehensively evaluate the performance of the proposed FedSDWC model, we selected two types of state-of-the-art (SOTA) FL models as baselines for comparison: (1) algorithms with OOD detection capabilities, including FedLN \citep{wei2022mitigating}, FOSTER \citep{yu2023turning}, and FedATOL \citep{zheng2023out}; (2) algorithms with OOD generalization capabilities, including FedT3A \citep{iwasawa2021test}, FedIIR  \citep{guo2023out}, FedTHE \citep{jiang2022test}, FOOGD \citep{liao2024foogd}, FedICON \citep{tan2023heterogeneity}, and PerAda \cite{xie2024perada}, FedCiR\citep{li2024fedcir}. Additionally, we introduced the classical heterogeneous FL method FedAvg \citep{mcmahan2017communication} as a baseline for comparison. Specifically, FedLN, FedATOL, and FedT3A are implementations that combine the LogitNorm, ATOL, and T3A algorithms with the FedAvg aggregation method, respectively. 

\textbf{Evaluation Metrics.} To evaluate the model's generalization ability on ID and OOD data, we report the accuracy on ID and ID-C data, denoted as \textbf{ID-Acc.} and \textbf{ID-C-Acc.}, respectively. Additionally, to assess the model's performance in OOD detection, we employ the maximum softmax probability (MSP) method \citep{hendrycks2016baseline} and use the area under the receiver operating characteristic curve (\textbf{AUROC}) and the false positive rate at 95\% true positive rate (\textbf{FPR95}) as evaluation metrics \citep{hendrycks2016baseline}. AUROC measures the model's ability to distinguish between ID and OOD data, with higher values indicating better performance, while FPR95 reflects the false positive rate when maintaining a high true positive rate (95\%), with lower values indicating better performance.

\textbf{Non-IID Client Data.}
To simulate the Non-IID nature of client data in real-world scenarios, this paper follows the method proposed by \cite{hsu2019measuring} to construct heterogeneous datasets. Specifically, we assume that for each client \(k\), its training samples follow a \(C\)-class categorical distribution parameterized by a class proportion vector \(\mathbf{q}_k\). This vector \(\mathbf{q}_k\) is sampled from a Dirichlet distribution, i.e., \(\mathbf{q}_k \sim \text{Dir}(\alpha \mathbf{p})\). Here, \(\mathbf{p}\) represents the prior distribution over the \(C\) classes; for instance, in datasets like CIFAR-10/100, \(\mathbf{p}\) is commonly set to a uniform distribution. The concentration parameter \(\alpha > 0\) controls the degree of data similarity (or heterogeneity) among clients. Smaller \(\alpha\) values lead to more pronounced data heterogeneity among clients (e.g., as \(\alpha \to 0\), each client might predominantly hold data from a single, randomly selected class). Conversely, larger \(\alpha\) values cause client data distributions to become more homogeneous and approach the global prior distribution \(\mathbf{p}\) (as \(\alpha \to \infty\), all client distributions converge to \(\mathbf{p})\). To generate each client's local dataset, we assign a corresponding number of samples for each class from the global training dataset to client \(k\) based on its sampled class proportion vector \(\mathbf{q}_k\).
Furthermore, to evaluate the model's generalization performance on unseen clients, we utilize the PACS domain generalization benchmark dataset. In this setup, we assign data from several source domains of the PACS dataset to three training clients, with each client holding data exclusively from one unique source domain. Data from the remaining one or more unseen (target) domains is then exclusively used to test the model's generalization ability and robustness against domain shift.

\textbf{Training details.} 
For the CIFAR-10 and CIFAR-100 datasets, we use the WideResNet \citep{zagoruyko2016wide} model.
For the Tiny TinyImageNet dataset, we use the ResNet-18 \citep{he2016deep} model. In the local training on the clients, we set 5 training epochs uniformly and use the SGD optimizer for model optimization. In all experiments, the number of clients is set at \( K = 10 \). Additionally, the learning rate for both the causal model and the feature extraction model is set to $0.001$. Our framework is implemented using PyTorch 1.10.0, and all training is performed on NVIDIA GeForce RTX 3090 GPUs.
To simulate the Non-IID scenario among clients, we sample data by label using a Dirichlet distribution, with the degree of Non-IID controlled by the parameter $\alpha$. A lower $\alpha$ value indicates greater heterogeneity in the data distribution across clients. In this work, we select $\alpha$ values of 0.1, 0.5, and 5. 

\setcounter{table}{0}   
\setcounter{figure}{0}
\setcounter{equation}{0}
\counterwithin{figure}{section}
\numberwithin{figure}{section}
\counterwithin{table}{section}
\numberwithin{table}{section}
\section{Extensive Experiment Results}\label{appendix:d}
\label{appendix:extensive}
\subsection{Supplementary experiment on the comparison of OOD generalization performance on Non-IID data.}

\begin{sidewaystable}[thp]
\centering
\caption{The comparison results of federated OOD generalization on Cifar-100 ($\alpha=0.1$).
}
\label{tab:gen:cifa100}
\begin{tabular}{cccccccccccc}
\toprule
\multirow{2}{*}{Corruption Type} & \multicolumn{9}{c}{Method}  \\
\cmidrule{2-12}
&FedAvg& FedLN & FOSTER& FedATOL  &FedT3A & FedIIR & FedTHE & FOOGD & PerAda & FedICON & Ours \\
\midrule
None &  51.67 & 52.48 & 72.54 & 43.65 & 51.67 &51.63 & 73.83 & 53.84 & 55.12 & 72.22 &  \textbf{78.05}\\
Brightness & 46.85 & 48.15 & 67.50 & 41.08 & 51.50 & 47.88 & 69.09 & 51.69 & 53.74  & 67.79 & \textbf{75.90}\\
Spatter & 42.41 & 43.90 & 63.32 & 36.39 & 47.32 & 42.04 & \textbf{64.11} & 49.59 & 47.26 & 62.92 & 62.61\\
Gaussian Blur & 34.18 & 36.29 & \textbf{55.41} & 30.63 & 38.98 & 35.39 & 54.56 & 40.63 & 40.10 &  54.41 & 41.98  \\
Saturate & 38.59 & 39.43 & 58.40 & 35.61 & 42.19 & 38.92 & 59.87 & 44.87 & 39.12 & 58.24 & \textbf{69.96} \\
Speckle Noise & 26.43 & 30.53 & 45.67 & 24.47 & 29.31 & 27.47 & 43.39 & 32.86 & 30.51 & 44.28 & \textbf{53.90} \\
Zoom Blur & 33.82 & 36.51 & 56.06 & 30.63 & 39.34 & 34.75 & 55.92 & 41.62 & 40.05 & 55.38 & \textbf{64.31} \\
Fog & 36.15 & 37.11 & 56.01 & 33.87 & 41.45 & 36.80 & 57.69 & 40.98 & 38.46 & 56.16 & \textbf{69.50}\\
Shot Noise & 26.37 & 30.28 & 44.96 & 24.01 & 29.20 & 27.03 & 42.33 & 32.81 & 31.02 & 43.64 & \textbf{55.36}\\
Frosted Glass Blur & 20.96 & 27.32 & 39.94 & 18.40 & 27.48 & 19.67 & 36.78 & 27.44 & 23.98 & 37.12 & \textbf{61.88}\\
Gaussian Noise & 21.21 & 24.83 & 38.43 & 19.67 & 22.50 & 21.79 & 35.51 & 28.28 & 30.25 & 37.00 & \textbf{49.97}\\
Motion Blur & 32.95 & 35.09 & 53.53 & 30.23 & 37.65 & 33.34 & 53.13 & 39.68 & 38.33 & 53.58 & \textbf{64.44}\\
Snow & 35.09 & 38.60 & 55.40 & 32.39 & 39.41 & 35.69 & 55.80 & 40.64 & 36.77 & 54.79 & \textbf{61.18}\\
Elastic Transform & 38.65 & 41.49 & 61.11 & 33.89 & 44.72 & 39.36 & 61.34 & 47.47 & 45.03 & 60.13 & \textbf{66.23}\\
Defouce Blur & 39.17 & 41.05 & 60.76 & 34.67 & 44.18 & 39.92 & 60.75 & 46.23 & 43.03 & 60.08 & \textbf{68.60}\\
Pixelate & 34.41 & 36.11 & 54.97 & 32.51 & 42.10 & 33.31 & 51.60 & 42.52 & 38.05 & 52.17 & \textbf{61.87}\\
Contrast & 26.39 & 27.10 & 42.91 & 26.96 & 29.88 & 26.94 & 44.98 & 30.98 & 33.12 & 43.62 & \textbf{61.97}\\
Frost & 32.53 & 35.38 & 52.36 & 29.50 & 37.40 & 33.33 & 53.15 & 38.54 & 41.98 & 51.08 & \textbf{61.88} \\
Impulse Noise & 22.99 & 24.26 & 40.68 & 21.58 & 23.65 & 21.84 & \textbf{38.44} & 26.24 & 30.19 & 39.21 & 37.61  \\
Jpeg Compression & 41.17 & 43.36 & 63.10 & 33.64 & 46.63 & 41.90 & \textbf{63.55} & 45.81 & 41.75 & 62.57 & 61.31 \\
\midrule
Avg. & 34.10 & 36.46 & 54.15 & 30.69 & 38.33 & 34.45 &  53.79 & 40.14 & 38.89 & 53.32 & \textbf{61.43}\\
\bottomrule
\end{tabular}
\end{sidewaystable}

Table \ref{tab:gen:cifa100} shows the experimental results of FedSDWC and different baseline methods on the CIFAR-100 dataset. In the highly heterogeneous data distribution setting (\(\alpha = 0.1\)), we introduced different types of data contamination. The results indicate that classical non-IID algorithms, such as FedAvg, exhibit weak generalization performance and struggle to effectively handle OOD data. FedIIR, based on invariant learning, aims to enhance the model's generalization ability by focusing on learning invariant features. However, its performance in the non-IID FL environment remains limited, especially under highly heterogeneous data distributions. Although FedIIR improves model stability to some extent, its adaptability to OOD samples is still insufficient. Additionally, other baseline methods show weaker generalization performance compared to our approach. Therefore, our experimental results demonstrate that FedSDWC, by further incorporating causal representations in both invariant and variable features, achieves SOTA performance across various distribution shift scenarios, further validating its effectiveness.

\begin{table}[ht]
\centering
\caption{Experimental results of federated OOD generalization and detection on TinyImageNet.}
\label{tab:gen:tinyimagenet}
\begin{tabular}{lcccc}
\toprule
Method & ID-Acc.\(\uparrow\) & ID-C-Acc.\(\uparrow\) & FPR95\(\downarrow\) & AUROC\(\uparrow\) \\
\midrule
FedAvg & 29.51 & 15.18 & 69.22 & 80.17 \\
FedLN & 38.23 & 15.64 &  61.40 & 82.31 \\
FedATOL & 23.32 & 13.69 & 29.04 & 94.91 \\
FedT3A & 29.46 & 00.50 & 69.14 & 80.08 \\
FedIIR & 38.01 & 14.90 & 78.84 & 69.38 \\
PerAda & 30.14 & 15.65 & 77.47 & 65.63 \\
Ours & \textbf{41.40} & \textbf{19.81} & \textbf{25.51} & \textbf{96.77} \\
\bottomrule
\end{tabular}
\end{table}

Table \ref{tab:gen:tinyimagenet} shows the performance of FedSDWC and different baseline methods on the TinyImageNet dataset. In the experiment, we use the brightness ACC as ID-C-Acc, which represents the OOD generalization performance. The results in the table indicate that, compared to all baseline methods, our approach not only achieves the best generalization ability but also delivers the best performance on in-distribution data. Notably, our model also achieved excellent results in OOD detection.

\begin{sidewaystable}[thp]
\centering
\caption{Experimental results of federated OOD generalization on different datasets.}
\label{tab:gen:diffalpha}
\begin{tabular}{lcccc|cccc}
\toprule
\multirow{4}{*}{Method} & \multicolumn{4}{c}{Cifar-10} & \multicolumn{4}{c}{Cifar-100} \\
\cmidrule(r){2-5} 
\cmidrule(r){6-9} 
& \multicolumn{2}{c}{$\alpha=0.5$} & \multicolumn{2}{c}{$\alpha=5$} & \multicolumn{2}{c}{$\alpha=0.5$} & \multicolumn{2}{c}{$\alpha=5$}\\
\cmidrule(r){2-3}
\cmidrule(r){4-5}
\cmidrule(r){6-7}
\cmidrule(r){8-9}
& ID-Acc & ID-C-Acc. & ID-Acc & ID-C-Acc. & ID-Acc & ID-C-Acc. & ID-Acc & ID-C-Acc.\\
\midrule
FedAvg & 86.59 & 83.72 & 86.50 & 85.08 & 58.28 & 54.62 & 61.40 & 56.72\\
FedLN & 86.10 & 84.20 &  87.20 & 85.08 & 59.39 & 53.86 & 61.00 & 56.33\\
FOSTER& 86.92 & 85.82 & 87.83 & 85.96 & 62.45 & 57.62 & 53.80 & 49.28\\
FedATOL & 87.55 & 85.64 & 89.27 & 88.28 & 60.62 & 56.63 & 64.16 & 63.61\\
FedT3A & 86.59 & 82.85 & 86.50 & 85.01 &  59.07 & 55.42 & 61.64& 55.51\\
FedIIR & 86.75 & 84.75 & 87.77 & 86.10 & 58.66 & 55.72 & 61.70 & 57.65\\
FedTHE & 89.14 & 87.68 & 88.14 & 86.18 & 66.22 & 61.19 & 61.03 & 57.03\\
FOOGD & 88.36 & 87.26 & 88.90 & 88.25 & 61.82 & 59.91 & \textbf{64.96} & 64.18\\
FedICON & 75.83 & 75.35 & 87.20 & 85.39 & 65.86 & 61.83 & 62.11 & 57.62\\
Ours & \textbf{92.05} & \textbf{91.08} & \textbf{91.15}& \textbf{91.75} & \textbf{67.89} & \textbf{63.68} & 63.64 & \textbf{66.82} \\
\bottomrule
\end{tabular}
\end{sidewaystable}

Table \ref{tab:gen:diffalpha} shows the experimental results of FedSDWC and other baseline methods on the CIFAR-10 and CIFAR-100 datasets under different \(\alpha\) values. The experiments use two metrics, ID-Acc. and ID-C-Acc., which represent the accuracy on in-distribution and out-of-distribution data, respectively. We use the brightness Acc as ID-C-Acc. 

On the CIFAR-10 dataset, when \(\alpha = 0.5\), our method (FedSDWC) performs excellently with ID-Acc. (92.05\%) and ID-C-Acc. (91.08\%), significantly outperforming other methods. Even under the \(\alpha = 5\) setting, our method still maintains high accuracy, with ID-Acc. at 91.15\% and ID-C-Acc. at 91.75\%, showing outstanding performance.

On the CIFAR-100 dataset, although the performance of other methods declines with both \(\alpha = 0.5\) and \(\alpha = 5\), our method still maintains high accuracy, especially when \(\alpha = 0.5\), where ID-Acc. is 67.89\% and ID-C-Acc. is 63.68\%, clearly outperforming the other baseline methods.

Overall, FedSDWC demonstrates strong OOD generalization ability across different data distribution settings and achieves the best performance on both in-distribution and out-of-distribution data.

\subsection{Supplementary experiment on the comparison of OOD detection performance on Non-IID data.}
\begin{table}[t]
\caption{Experimental results of federated OOD detection on Cifar-100.}
\label{tab:detection:cifar100}
\vskip 0.15in
\begin{center}
\begin{small}
\begin{sc}
\begin{tabular}{lcccccc}
\toprule
\multirow{2}{*}{Method} & \multicolumn{2}{c}{$\alpha=5.0$} & \multicolumn{2}{c}{$\alpha=0.5$} & \multicolumn{2}{c}{$\alpha=0.1$}\\
\cmidrule(r){2-3}
\cmidrule(r){4-5}
\cmidrule(r){6-7}
& FPR95$\downarrow$ & AUROC $\uparrow$ & FPR95$\downarrow$ & AUROC $\uparrow$& FPR95$\downarrow$ & AUROC $\uparrow$\\
\midrule
FedAvg & 72.68 & 70.59 & 72.84 & 70.86 & 78.35 & 67.16 \\
FedLN  & 69.18 & 75.87 & 68.31 & 73.41 & 66.94 & 74.82\\
FOSTER & 76.94 & 65.47 & 73.26 & 68.71 & 61.25 & 75.44\\
FedATOL& 80.27 & 60.51 & 70.10 & 79.27 & 65.26 & \textbf{81.64}\\
FedT3A & 72.77 & 70.44 & 72.86 & 70.88 & 78.36 & 67.22\\
FedIIR & 72.57 & 69.07 & 77.62 & 65.87 & 81.91 & 63.99\\
FedTHE & 71.43 & 69.01 & 72.95 & 69.38 & 64.73 & 75.16\\
FedICON& 70.91 & 70.84 & 69.99 & 71.03 & 61.36 & 77.12\\
Ours   & \textbf{58.88} & \textbf{79.45} & \textbf{60.56} & \textbf{79.80} & \textbf{61.09}& 79.23 \\
\bottomrule
\end{tabular}
\end{sc}
\end{small}
\end{center}
\end{table}

Table \ref{tab:detection:cifar100} shows the federated OOD detection experimental results on the CIFAR-100 dataset under different \(\alpha\) values. The evaluation metrics used are FPR95 and AUROC, where lower FPR95 and higher AUROC are better.

In the \(\alpha = 5.0\) setting, our method performs excellently with FPR95 (58.88) and AUROC (79.45), significantly outperforming other baseline methods. Even under the \(\alpha = 0.5\) and \(\alpha = 0.1\) settings, our method still maintains excellent performance. When \(\alpha = 0.5\), FPR95 is 60.56 and AUROC is 79.80.

Overall, FedSDWC demonstrates strong OOD detection capabilities across different \(\alpha\) value settings, achieving the best results in both FPR95 and AUROC.

Table \ref{fig:app:detection-cifar10} shows the federated OOD detection experimental results on the CIFAR-10 dataset under different \(\alpha\) values. 

In the \(\alpha = 0.5\) setting, our method performs excellently with FPR95 (30.72) and AUROC (90.85), significantly outperforming other baseline methods. In the \(\alpha = 5\) setting, our method also performs outstandingly, with FPR95 (29.65) and AUROC (91.32), clearly surpassing other methods.

Overall, FedSDWC demonstrates strong OOD detection capabilities across different datasets and \(\alpha\) value settings, achieving the best results in both FPR95 and AUROC.

\begin{table}[ht]
\caption{Main results of federated OOD detection on Cifar-10.}
\label{fig:app:detection-cifar10}
\vskip 0.15in
\begin{center}
\begin{tabular}{lcccc}
\toprule
\multirow{2}{*}{Method} & \multicolumn{2}{c}{$\alpha=0.5$} & \multicolumn{2}{c}{$\alpha=5$}\\
\cmidrule(r){2-3}
\cmidrule(r){4-5}
& FPR95$\downarrow$ & AUROC $\uparrow$ & FPR95$\downarrow$ & AUROC $\uparrow$\\
\midrule
FedAvg  & 43.70 & 84.18 & 38.24 & 85.37 \\
FedLN  & 39.26 & 89.64 & 33.33 & 90.87 \\
FOSTER & 42.03 & 83.91 & 36.42 & 86.19\\
FedT3A & 43.70 & 84.18 & 38.24 & 85.37 \\
FedIIR & 40.91 & 84.94 & 34.69 & 87.66\\
FedTHE & 40.28 & 85.30 & 35.35 & 86.79\\
FedICON & 56.19 & 79.88 & 35.63 & 86.45\\
Ours & \textbf{30.72} & \textbf{90.85} & \textbf{29.65} & \textbf{91.32}\\
\bottomrule
\end{tabular}
\end{center}
\end{table}

\subsection{Client Generalization on PACS Dataset.} To validate the effectiveness of FedSDWC in domain generalization tasks, this study adopts a leave-one-out training strategy, constructing the experimental environment with reference to the experimental setups of FedIIR\cite{guo2023out} and FOOGD \cite{liao2024foogd} (each client contains only single-domain data). All comparative models are pre-trained from scratch and strictly follow the adaptation methods of the original papers to ensure fairness. As shown in Table \ref{fig:app:generalization-pacs}, in the most challenging task of "excluding the Sketch domain," FedSDWC significantly outperforms existing methods.

\begin{table}[!ht]
\centering
\caption{OOD generalization for PACS dataset}
\label{fig:app:generalization-pacs}
\vskip 0.2in
\begin{tabular}{lccccc}
\toprule
Method & Art Painting & Cartoon & Photo & Sketch & Avg.\\
\midrule
FedAvg  & 97.21 & 62.58 & 91.00 & 35.28 & 71.52  \\
FedT3A & 97.13 & 75.71& 93.21 & 37.40 & 75.86 \\
FedIIR & 86.86 & 80.29 & 88.98 & 31.38 & 71.88 \\
FedTHE & 96.17 & 90.72 & 93.57 & 29.14 & 77.40\\
FOOGD & 97.46 & 89.32 & 91.48 & 41.40 & 79.92\\
Ours & \textbf{98.56} & \textbf{93.42} & \textbf{96.06} & \textbf{42.24} & \textbf{82.57} \\
\bottomrule
\end{tabular}
\end{table}

\subsection{Ablation Study}
\begin{figure}[ht]
\begin{center}
\subfigure[FedAvg]{
\begin{minipage}{0.49\columnwidth}
    \centering
    \includegraphics[width=1\linewidth]{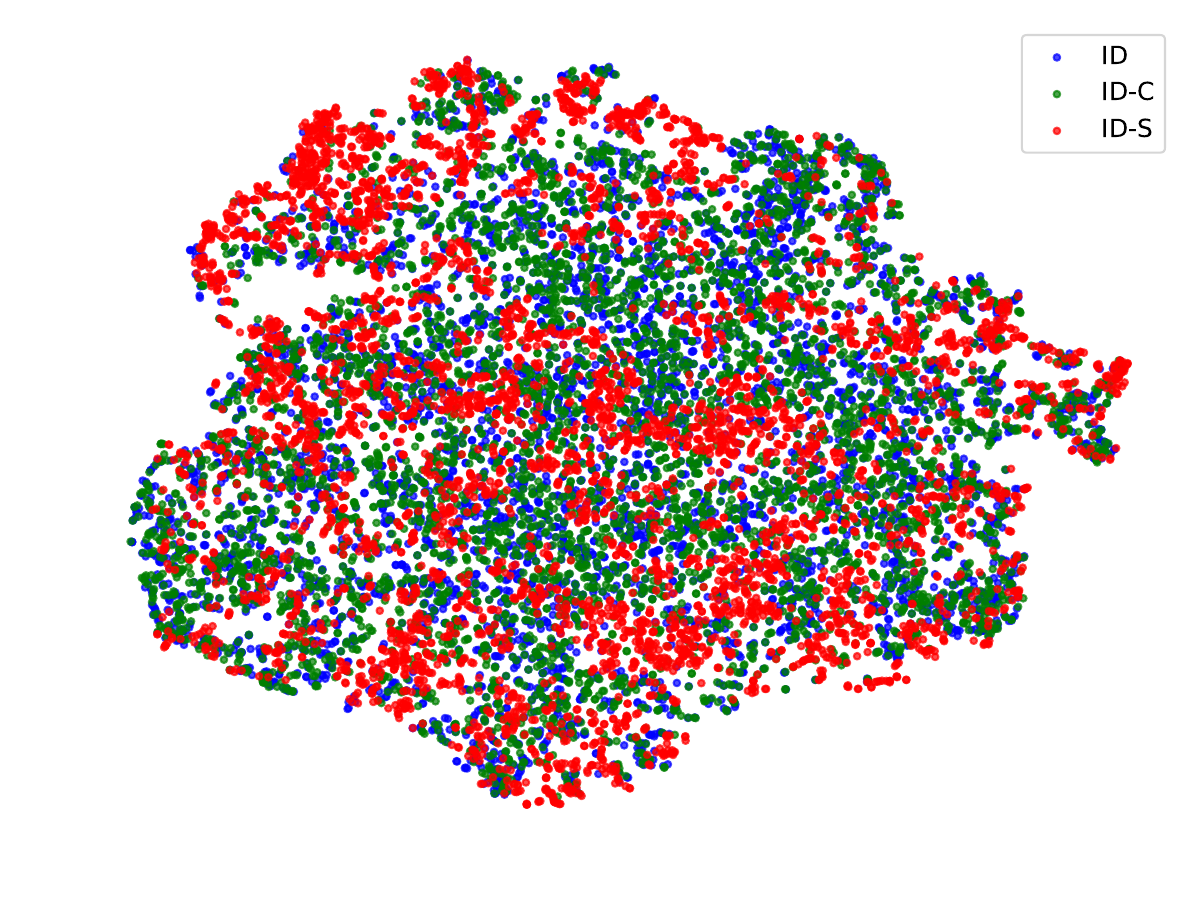}
\end{minipage}
}
\subfigure[PerAda]{\begin{minipage}{0.49\columnwidth}
    \centering
    \includegraphics[width=1\linewidth]{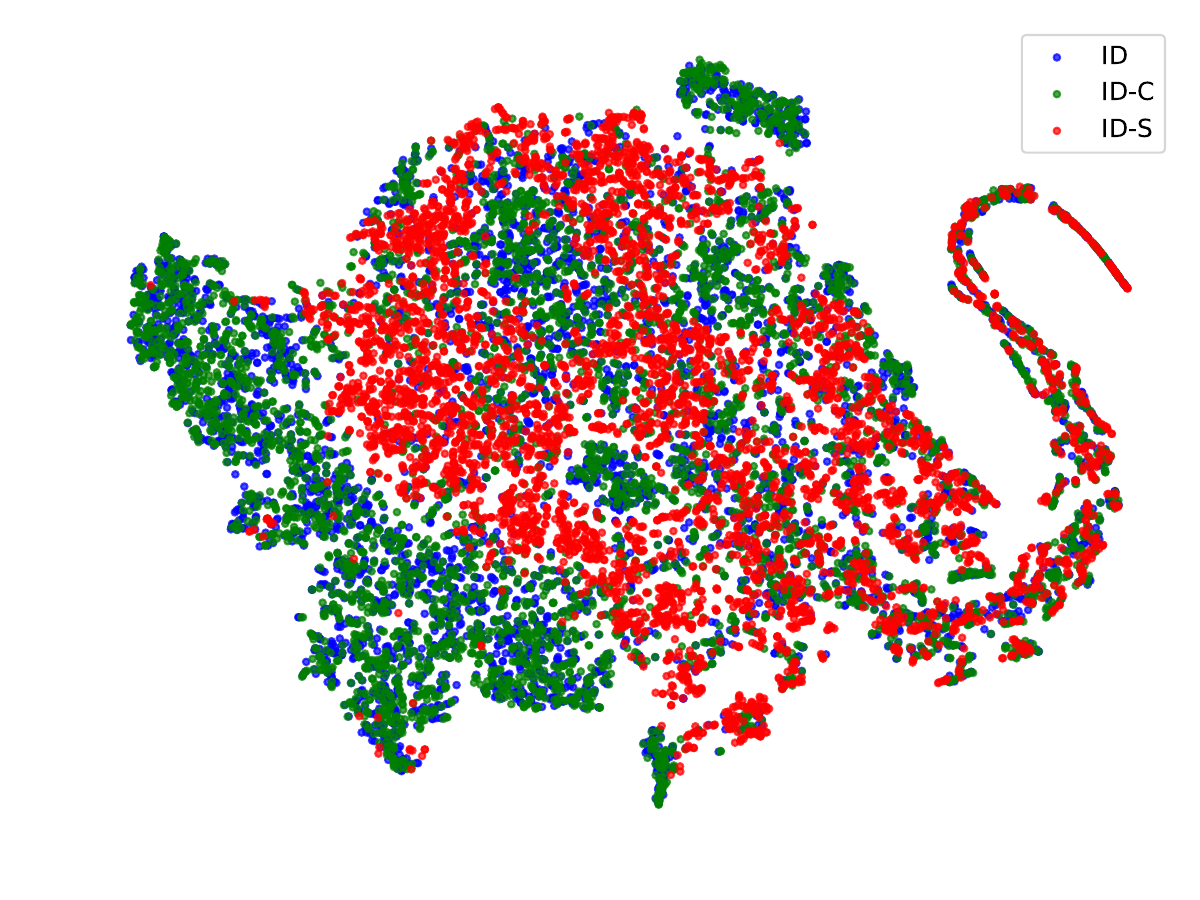}
\end{minipage}}

\subfigure[FedIIR]{
\begin{minipage}{0.48\columnwidth}
    \centering
    \includegraphics[width=1\linewidth]{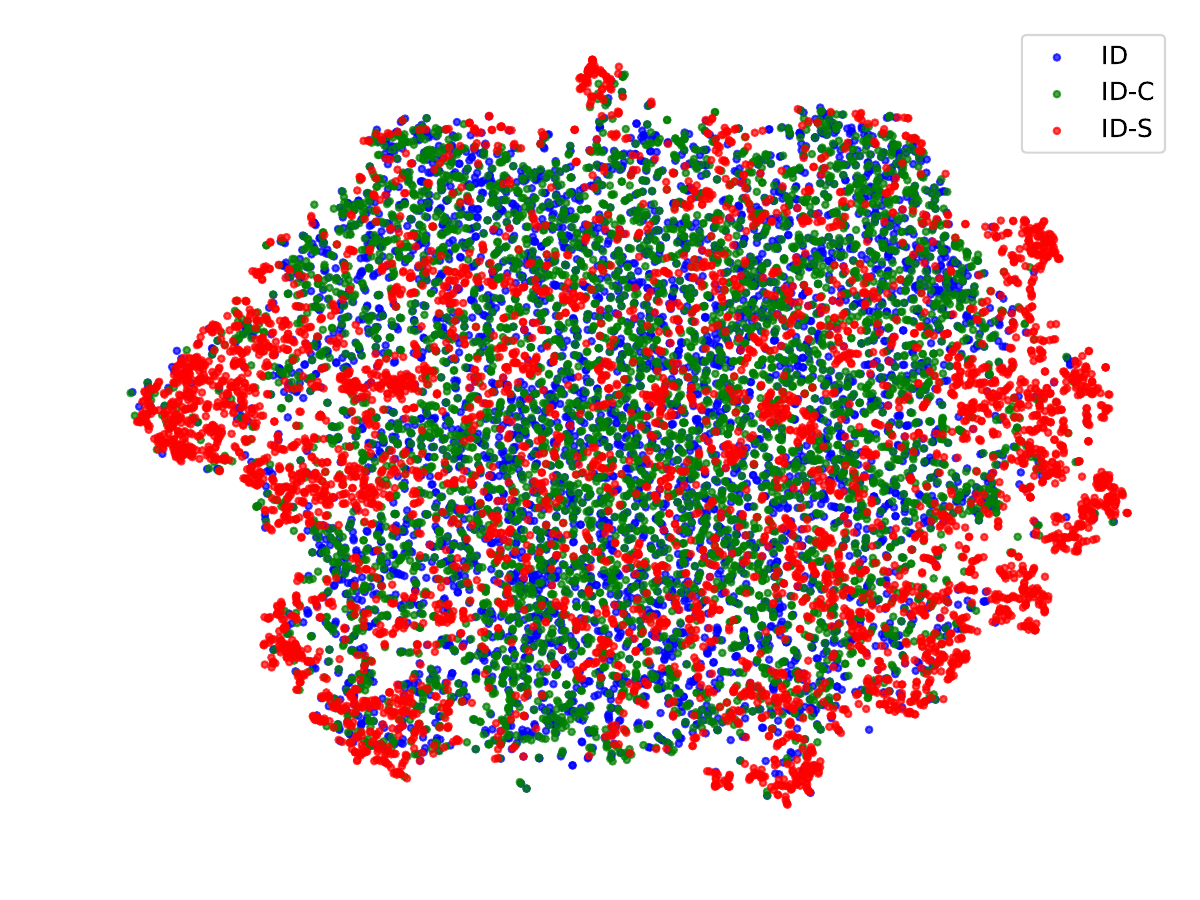}
\end{minipage}
}
\subfigure[Ours]{
\begin{minipage}{0.48\columnwidth}
    \centering
    \includegraphics[width=1\linewidth]{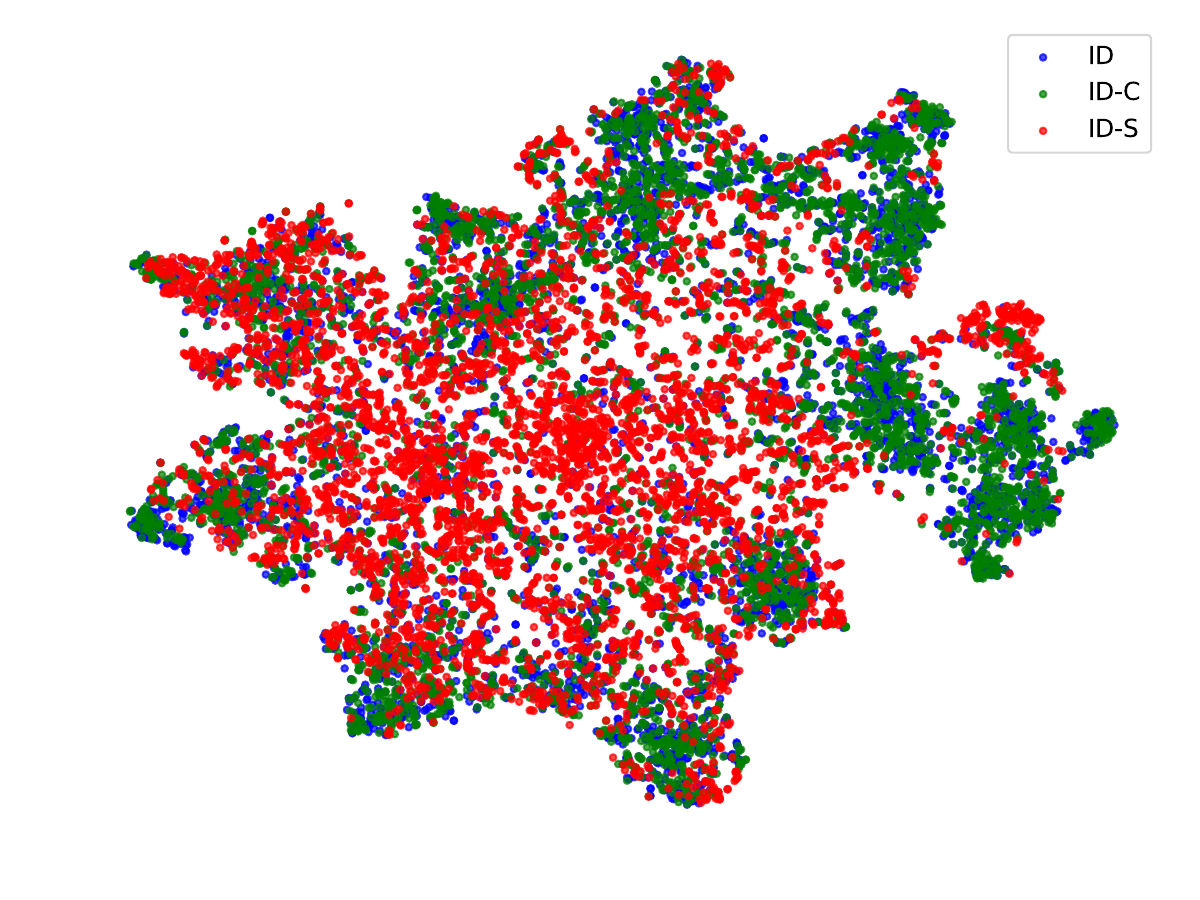}
\end{minipage}
}
\caption{T-SNE visualization of different models ($\alpha=0.1$).}
\label{fig:t-sne:alpha01}
\end{center}
\end{figure}

\textbf{Visualization.} To further explore the data distribution characteristics in federated learning OOD generalization/detection models, we use t-SNE to visualize the data representations under stronger heterogeneity. The experiment is based on the CIFAR-10 dataset (\(\alpha=0.1\)), comparing FedAvg, PerAda, FedIIR, and our FedSDWC model. From the Figure \ref{fig:t-sne:alpha01}, it can be observed that only our model performs category partitioning in the ID data, with tighter distributions between categories. Additionally, our model constructs a clearer decision boundary between ID data and ID-S data compared to other baseline methods.

\begin{figure*}[ht]
\vskip 0.2in
\begin{center}
\subfigure[ID-Acc.]{
\begin{minipage}{0.6\columnwidth}
    \centering
    \includegraphics[width=1\linewidth]{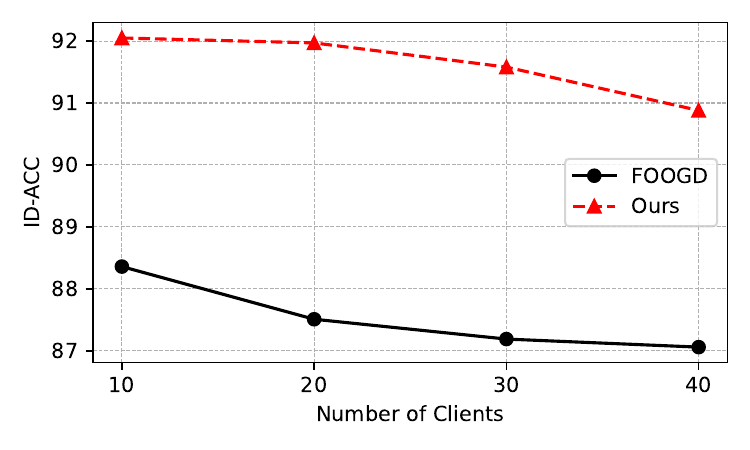}
\end{minipage}
}
\subfigure[ID-C-Acc.]{\begin{minipage}{0.6\columnwidth}
    \centering
    \includegraphics[width=1\linewidth]{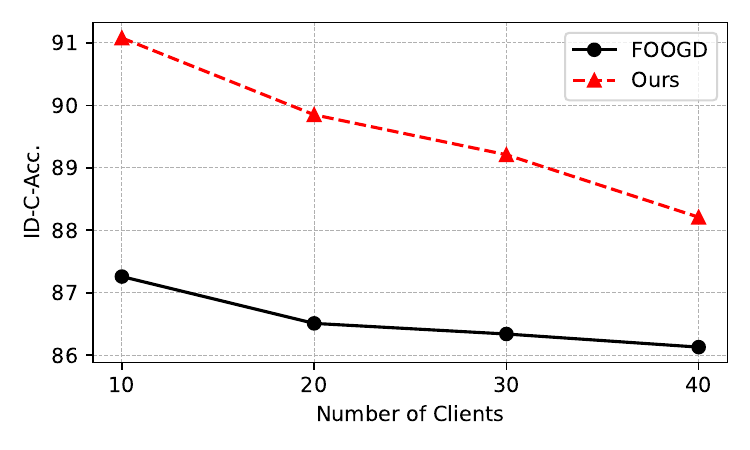}
\end{minipage}}

\caption{Effect of participating clients numbers $K$.}
\label{fig:Keffect}
\end{center}
\end{figure*}

\textbf{Impact of client number $K$ on generalization performance.} To investigate the effect of varying client scales on model performance, we conducted experiments under the settings of $K={10,20,30,40}$. As shown in Figure \ref{fig:Keffect}, comparative experiments with the SOTA baseline method FOOGD demonstrate that the proposed algorithm consistently achieves superior generalization performance on different scales of participation of clients.

\begin{figure}[ht]
    \centering
    \includegraphics[width=0.6\linewidth]{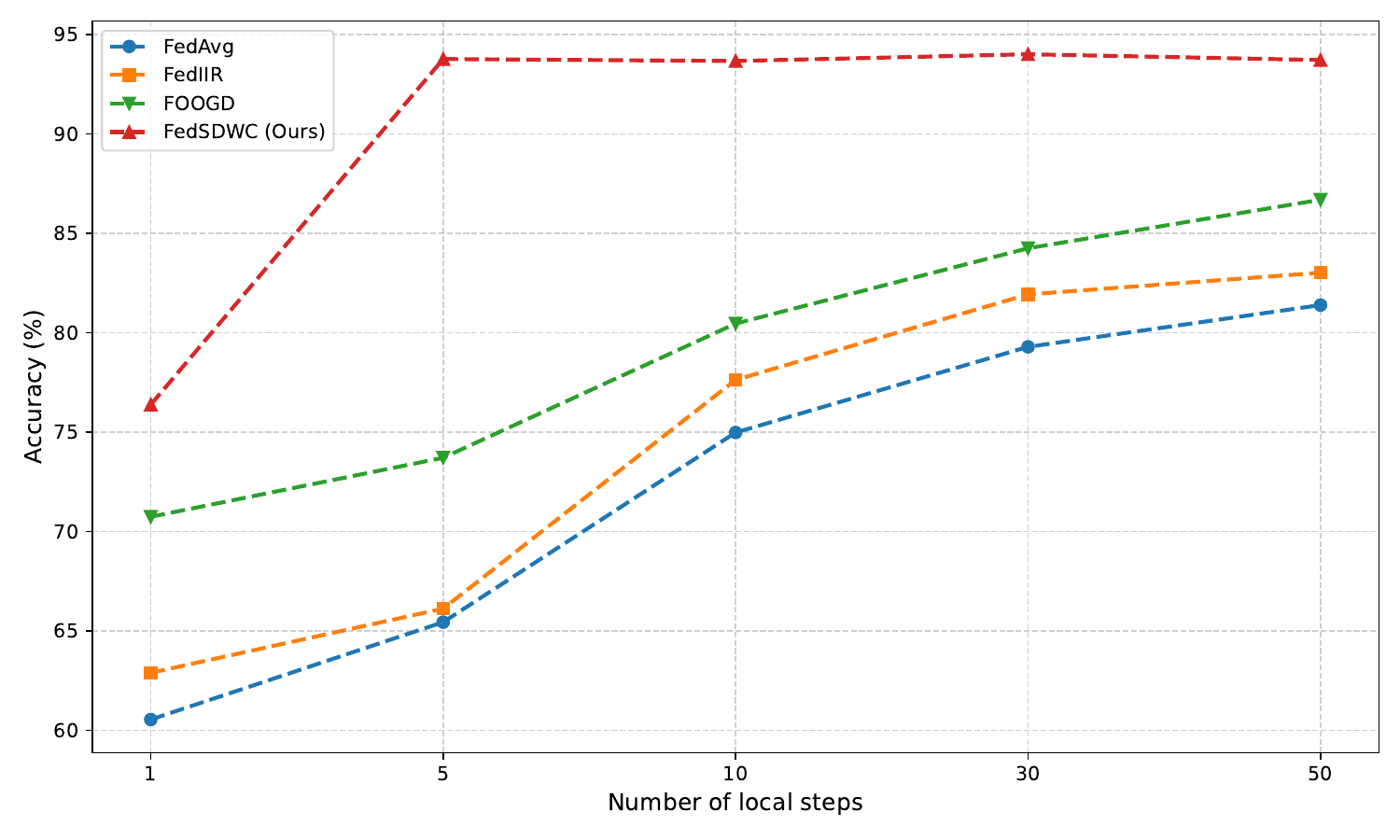}
    \caption{ Test accuracy (\%) of different methods with
varying local updating steps on Cifar-10-C.}
    \label{fig:localEpoch}
\end{figure}

\textbf{Impact of local steps:} We investigate the impact of the number of local update steps on model generalization performance, evaluating the accuracy of different methods on the Cifar-10-C dataset. As shown in Figure \ref{fig:localEpoch}, appropriately increasing the number of local update steps improves the accuracy of all methods. Notably, FedSDWC consistently outperforms the other methods across all settings, with especially strong performance when the number of local steps is small (e.g., 1 and 5), demonstrating its significant advantage in low communication cost scenarios.

\subsection{Stability Analysis of Various Methods under Different Corruption Types} 
This study aims to systematically evaluate the performance stability of various federated learning methods under different data corruption conditions. The experimental design encompasses 20 common types of data corruption (e.g., brightness variations, noise interference, blurring, etc.), and comprehensively tests the generalization capabilities of 11 federated learning algorithms (including baseline methods such as FEDAVG and FEDTHE, as well as our proposed method) under these corruptions.  


\begin{figure}[!ht]
    \centering
    \includegraphics[width=1\linewidth]{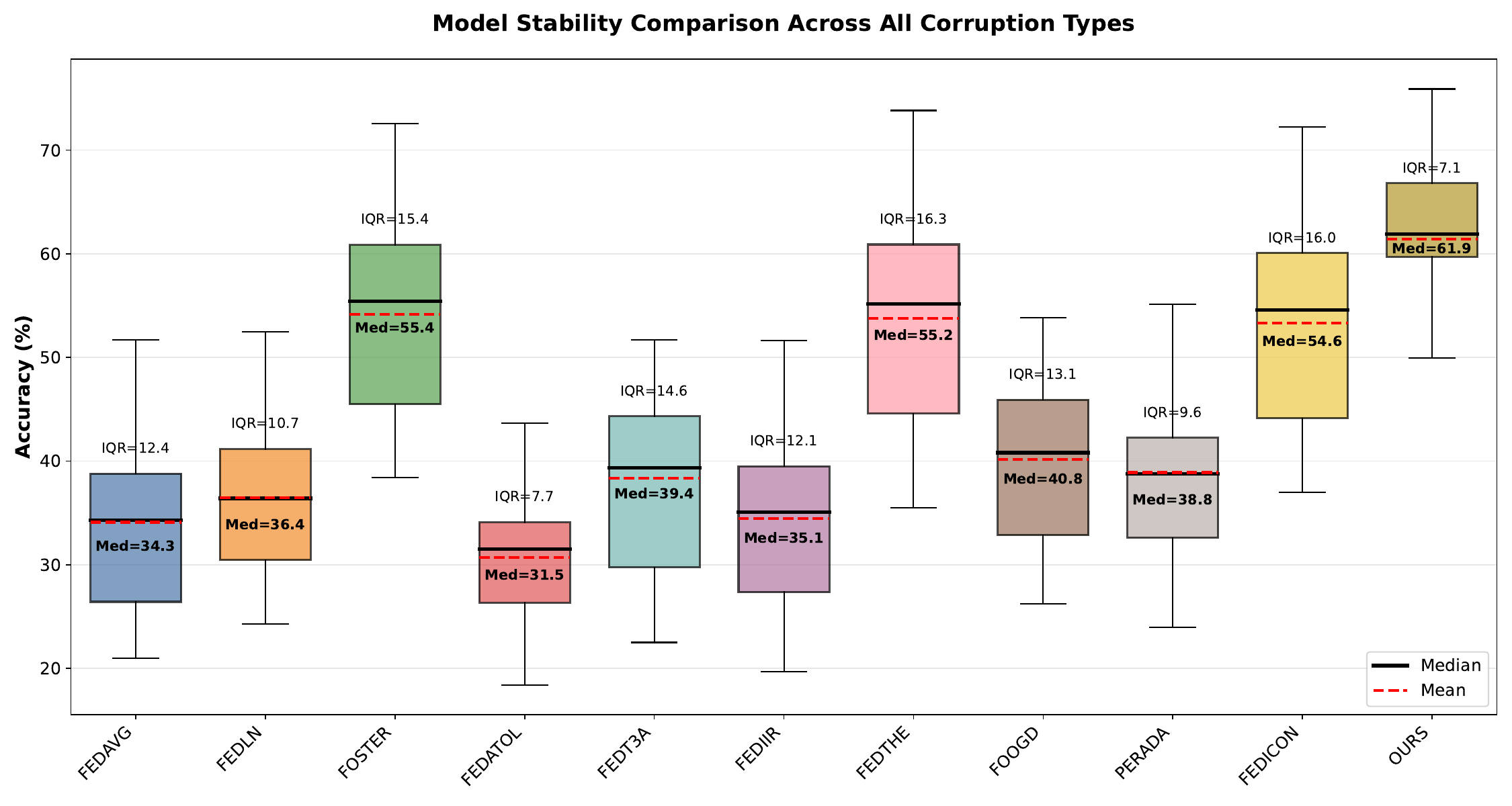}
    \caption{Comparison of Anti-Corruption Performance Stability Across Methods on CIFAR-100-C.}
    \label{fig:cifar100_box_plot}
\end{figure}
\textbf{Performance on CIFAR-100-C (as shown in Figure \ref{fig:cifar100_box_plot}):}
For the more complex CIFAR-100-C dataset, the our method also delivers outstanding performance. It achieves a median accuracy of 61.9\%, significantly outperforming most baseline methods (e.g., FEDAVG's median accuracy is 34.3\%). Our method's IQR is 7.1, substantially smaller than those of traditional methods (e.g., FEDAVG's IQR of 12.4 and FEDICON's IQR of 16.0). This further confirms that the our method maintains higher consistency and stability across various corruption conditions. The boxplots visually illustrate the advantages of our method—its box range is more compact (reflecting a smaller IQR), its median position is higher, and it generally exhibits fewer outliers. This indicates that even when facing challenging corruption types such as Frosted Glass Blur and Impulse Noise (both included in the overall evaluation of 20 corruption types), the our method can still maintain robust performance.

In summary, the series of experimental results on both CIFAR-10-C and CIFAR-100-C datasets strongly demonstrate the superiority of our method in handling diverse data corruptions. Not only does it achieve leading classification accuracy, but more importantly, it exhibits stronger stability (i.e., smaller IQR). This makes it particularly valuable for processing complex and variable real-world data, showcasing significant practical utility and technical advantages.

\section{Related Works}\label{sec:rel}
\subsection{Federated Learning with Non-IID data}
FL has gained widespread attention in recent years due to its critical role in data privacy protection scenarios. However, the significant heterogeneity in non-IID data across clients poses a major challenge to the performance of the global model. Classic aggregation algorithms, such as FedAvg \citep{mcmahan2017communication} , train the global model by averaging parameter updates from clients. However, its performance often suffers significantly in non-IID scenarios. To address this issue, FedProx \citep{li2020federated} introduces a regularization term in the objective function to constrain the deviation between local and global models, alleviating inconsistencies caused by non-IID data. Additionally, FRAug \citep{chen2023fraug} employs representation augmentation techniques by sharing a latent generator to capture consistency across clients, thereby improving model performance.
Meanwhile, personalized federated learning (PFL) has emerged as another critical approach to tackling data heterogeneity challenges. For example, FedL2P \citep{lee2024fedl2p} leverages a meta-network to learn each client's BatchNorm and learning rate parameters, generating personalized strategies for clients based on local data statistics to achieve better adaptability. Similarly, pFedBreD \citep{shi2024prior} enhances model flexibility and adaptability in personalized scenarios by decoupling personalized priors from local objective functions using regularized Bregman divergence. While these methods effectively improve model performance under heterogeneous data distributions, they often lack mechanisms to handle OOD data, leading to suboptimal generalization performance.

\subsection{OOD Generalization}
\textbf{OOD generalization} refers to extracting invariant feature-label relationships in the presence of covariate shift to ensure the robust deployment of models in real-world scenarios \citep{lv2023duet, liao2024foogd, mildner2025federated, nguyen2025federated}. Currently, OOD generalization research primarily focuses on three approaches: invariant learning \citep{arjovsky2019invariant, guo2023out, li2022learning}, disentangled learning \citep{kong2022partial, bai2024diprompt}, and causal inference \citep{gui2024joint}. Invariant learning, proposed by \cite{arjovsky2019invariant}, employs strategies such as Invariant Risk Minimization (IRM) to optimize feature representations across different covariate shift scenarios, extracting invariant features with strong generalization capabilities. Disentangled learning aims to separate different semantic factors within the data, extracting stable factors related to the labels to construct more robust feature representations. Causal inference methods, based on Structural Causal Models (SCM), infer latent factors within the data to eliminate spurious correlations and improve model generalization. These approaches collectively strive to enhance the adaptability and reliability of models when dealing with out-of-distribution data.

\subsection{OOD Detection}
\textbf{OOD Detection} aims to distinguish between ID and OOD samples during the inference phase to identify unknown data. Common OOD detection methods can be categorized into three types: classification-based \citep{djurisic2023extremely, linderman2023fine, park2023nearest}, distance-based \citep{sun2022out, galesso2023far, ming2024does}, and density-based \citep{wang2022vim, yang2023full} approaches. Classification-based methods typically leverage the neural network's output, using the maximum softmax probability as an ID sample discrimination metric, focusing on improving detection performance through enhanced model outputs \citep{hendrycks2016baseline}. Distance-based methods rely on the relative distance between the sample and known class centroids or prototypes \citep{yang2024generalized}. For example, \cite{lee2018simple} proposed using the minimum Mahalanobis distance for detection, assuming that OOD samples should be farther from the ID distribution. 
Density-based methods, on the other hand, construct probabilistic models to characterize known data distributions and label samples in low-density regions as OOD. Common approaches include flow-based methods \citep{jiang2021revisiting} and variational autoencoders \citep{xiao2020likelihood, wu2023discriminating}.